\documentclass[sigconf]{acmart}

\renewcommand\footnotetextcopyrightpermission[1]{} % removes footnote with conference information in first column
\usepackage{algorithmic}
\usepackage{graphicx}
\usepackage{textcomp}
\usepackage{xcolor}
\usepackage{balance}
\usepackage{listings}
\usepackage{enumitem}
\usepackage{array}
\usepackage{subcaption}

\definecolor{codegreen}{rgb}{0,0.6,0}
\definecolor{codegray}{rgb}{0.5,0.5,0.5}
\definecolor{codepurple}{rgb}{0.58,0,0.82}
\definecolor{backcolour}{rgb}{0.95,0.95,0.92}
\lstdefinestyle{code_style}{
    backgroundcolor=\color{backcolour},
    commentstyle=\color{codegreen},
    keywordstyle=\color{magenta},
    numberstyle=\tiny\color{codegray},
    stringstyle=\color{codepurple},
    basicstyle=\ttfamily\footnotesize,
    breakatwhitespace=false,         
    breaklines=true,                 
    captionpos=b,                    
    keepspaces=true,                 
    numbers=left,                    
    numbersep=5pt,                  
    showspaces=false,                
    showstringspaces=false,
    showtabs=false,                  
    tabsize=2
}

\newcolumntype{P}[1]{>{\centering\arraybackslash}m{#1}}

\AtBeginDocument{%
  }

\setcopyright{none}
\copyrightyear{2025}
\acmYear{2025}
\acmDOI{XXXXXXX.XXXXXXX}
\acmConference[ArXiv Preprint]{ArXiv Preprint}{2025}{USA}

\acmBooktitle{ArXiv Preprint, 2025, Santa Fe, NM, USA}

\acmSubmissionID{3324}

\begin{document}

%\title{Coupling AI and Non-AI Agents Using PDES for Scalable and Trustworthy Simulations \atanu{SHOULD WE CHANGE THE TITLE?}}
\title{Scalable, Symbiotic, AI and Non-AI Agent Based Parallel Discrete Event Simulations}

\author{Atanu Barai}
\affiliation{%
  \institution{Los Alamos National Laboratory}
  \city{Los Alamos}
  \state{NM}
  \country{USA}
  }
\email{abarai@lanl.gov}

\author{Stephan Eidenbenz}
\affiliation{%
  \institution{Los Alamos National Laboratory}
  \city{Los Alamos}
  \state{NM}
  \country{USA}
  }
\email{eidenben@lanl.gov}

\author{Nandakishore Santhi}
\affiliation{%
  \institution{Los Alamos National Laboratory}
  \city{Los Alamos}
  \state{NM}
  \country{USA}
  }
\email{nsanthi@lanl.gov}


\begin{abstract}
To fully leverage the potential of artificial intelligence (AI) systems in a trustworthy manner, it is often desirable to couple multiple AI and non-AI systems together seamlessly for constraining and ensuring correctness of the output. This paper introduces a novel parallel discrete event simulation (PDES) based symbiotic methodology to combine multiple AI and non-AI agents in a causal, rule-based way. Our approach tightly integrates the concept of {\em passage of time}, with each agent considered as an \emph{entity} in the PDES framework and responding to prior requests from other agents. Such a coupling mechanism enables the agents to work in a co-operative environment towards a common goal while many tasks run in parallel throughout the simulation. It further enables setting up boundaries to the outputs of the AI agents by applying necessary dynamic constraints using non-AI agents while allowing for scalability through deployment of hundreds of such agents in a larger compute cluster. Distributing smaller AI agents can potentially enable extremely scalable simulations in the future, while also circumventing local memory bottlenecks for model parameter storage.

Within a parallel discrete event simulation involving both AI and non-AI agents, we break down the problem at hand into structured steps, when necessary, providing a set of multiple choices to the AI agents, and then progressively solve these steps towards a final goal. At each step, the non-AI agents act as unbiased auditors, examining and verifying each action by the AI agents so that certain rules of engagement are followed. We evaluate our approach by solving four problems from four different domains and comparing the results with those from AI models alone. Our results show that by adopting the coupling methodology, we can achieve greater accuracy in solving problems from various domains where the AI models struggle to solve the problems solely by themselves. Results show that overall accuracy of our approach is 68\% where as the accuracy of vanilla models is less than 23\%.
\end{abstract}



\maketitle

\section{Introduction}
\label{sec:intro}
Artificial intelligence (AI) is revolutionizing countless industries and enabling new discoveries in every field including science, energy, economics, and security thus transforming how we innovate. AI systems, when specifically trained on various topics, are capable of tasks once thought to be possible only by humans. However, a majority of AI models today run in isolation, unable to fully harness the potential of collective knowledge and parallel synergy, thus hindering further innovation by achieving shared objectives.

The growing complexity of pressing problems in science and other domains demands solutions that not only leverage the strengths of multiple AI systems but also of non-AI systems, all coupled together. While AI agents excel in tasks such as repetitive data processing, pattern recognition, content generation, and adaptive decision making, non-AI agents such as human experts, and rule-based deterministic/Monte Carlo systems, often possess domain-specific knowledge, practical constraints, and ethical/societal considerations that the AI systems can't yet fully encapsulate. Thus their collaboration can lead to more versatile, resilient, and context-aware systems capable of addressing a broad spectrum of challenges by pooling knowledge, sharing information, and coordinating actions that would be beyond the reach of any individual agent. Thus, these types of systems promise enhanced problem solving capabilities and greater adaptability to complex environments including autonomous transportation, drug discovery, education etc.

However, developing a robust coupling methodology to enable multi-agent AI systems is challenging due to several factors. Issues such as communication methodologies, timing/coordination among the agents, conflict resolution, handling biases, security considerations, and most importantly extreme scalability to handle a very large number of agents must be addressed to ensure an effective collaboration strategy.

This paper introduces a robust parallel discrete event simulation (PDES) based methodology to couple AI and non-AI agents in a rule-based way into cohesive multi-agent systems. Using PDES for coupling inherently addresses many of the problems mentioned previously. This strategy is capable of solving a problem by breaking it into many smaller, more manageable steps, involving both AI and non-AI agents to solve each stage in a causal event-driven manner. We use the open-source \emph{Simian}~\cite{simian} PDES engine to couple the agents. \emph{Simian} is simple and easy to use, yet powerful enough to support the deployment of thousands of such agents through a user-transparent parallelization mechanism which effectively uses the message passing interface (MPI) library. Each simulation agent is implemented as an \emph{entity} object of the Simian engine and is assigned to one of the MPI ranks for parallel processing. The \emph{entities} serve requests through a \emph{reqService} process which schedules a future event at the corresponding \emph{entity} to be processed by a user-defined \emph{event\ handler}.

The ability to couple and deploy hundreds of AI and non-AI agents also addresses another contemporary challenge faced by the AI community. Today's AI models utilize the extremely parallel computing power of graphics processing units (GPUs). With the introduction and rapid adoption of large language models (LLMs)~\cite{Bommasani2021FoundationModels,wei2022emergent_llm} such as Llama3~\cite{dubey2024llama3} and GPT4~\cite{openai2024gpt4},  computing requirements and demand for powerful GPUs have skyrocketed~\cite{strati-gpu-shortage-cross-region}, since these models are trained on large amounts of web data. Running local inference on these models can be costly and out of reach for a vast majority of researchers due to the requirement for very large compute and memory throughputs. By using our PDES based methodology, we are able to deploy thousands of smaller AI agents coupled with many non-AI agents to work together to solve a particular problem.

To demonstrate our approach more thoroughly, we solve four common problems from geometry, combinatorics, arithmetics and, graph-theory by coupling small language model (SLM) based AI agent with non-AI agents, where a small AI agent by itself is unable to solve these problems correctly. We consider models with 16B parameters or less as SLM, while anything larger is considered an LLM. This is based on current technology considerations, where 16B or lower models can be accommodated on low to medium scale consumer-grade GPUs from major vendors, albeit by employing suitable parameter quantization strategies. The general symbiotic approach to solving the problems is the following:
\begin{enumerate}[label=(\arabic*)]
\item We break down the problem into multiple smaller steps, 
\item SLM agents are asked in parallel to solve a step, 
\item Constraints in the solution space are applied by providing SLM agents with some pre-defined options to choose from, 
\item Non-SLM agents collate and then inspect SLM responses and provide feedback in the form of further prompts to the SLM agents, 
\item The whole process runs under the umbrella of a parallel discrete event simulation.
\end{enumerate}

These problems sometimes can be solved by a large language model (LLM) acting on its own. But most large models require GPU acceleration with very large local memory to run which is in turn very expensive and often scarce. In contrast, in our approach, we use SLMs that require far less powerful GPUs with low local memory requirements, and can even be accelerated on GPUs/custom-accelerators/NPUs integrated on the CPU-fabric. On the other hand, our method can optionally couple together multiple SLMs to get more accurate responses~\cite{shen2024collab_llms}. Our approach also has a simple verification step built in as the non-AI agent checks individual SLM answers, thus making the entire system much more trustworthy than a pure SLM/LLM approach.

The rest of the paper is organized as follows: Section~\ref{sec:related-works} summarizes the relevant related works and sets this work apart from those. Section~\ref{sec:background} covers the necessary background concepts used in this paper. Section~\ref{sec:methodology} provides an overview of our coupling methodology and shows how one can solve some common problems encountered in different domains by incorporating both AI and non-AI agents through PDES. Section~\ref{sec:results} goes over some experimental results. It compares the outcome of our coupled AI system with a traditional stand-alone AI system.  Finally, Section~\ref{sec:conclusion} concludes the paper.

%%%%%%%%%%%%%%%%%%%%%%%%%%%%%%%%%%%%%%%%%%%%%%%%%%%%%%%%%%%%
%%%%%%%%%%%%%%%%%%%%%%%%%%%%%%%%%%%%%%%%%%%%%%%%%%%%%%%%%%%%
%%%%%%%%%%%%%%%%%%%%%%%%%%%%%%%%%%%%%%%%%%%%%%%%%%%%%%%%%%%%
\section{Related Works}
\label{sec:related-works}
Coupling several models trained on varied topics into multi-agent systems has been an active field of research to improve the overall quality of AI model responses. The interplay of multiple agents require seamless integration of heterogenous agents bringing unique challenges regarding communication, coordination and interaction-rules. Wooldridge discussed various aspects of multi-agent systems including architectures and applications~\cite{wooldridge2009multiagent}. Researchers have also extensively studied those challenges, opportunities and limitations of integrating AI and non-AI entities~\cite{stone2000multiagent,alonso2002multiagentlearning,hunter2014measuring} including multi-agent deep-reinforcement learning~\cite{hernandez2019muti-reinforce}.

Baratta \emph{et al. }~\cite{BARATTA2023surv-human-robot} showed how AI-based robots can collaborate with human agents in industrial settings through continuous learning of human behavior. Ming \emph{et al. }~\cite{ming2023exploration} explored a framework for bridging AI based adaptive algorithms with deterministic non-AI control software within the context of autonomous vehicles.

In terms of language models, the coupling approaches vary depending on model architecture and the underlying learning settings. Shen \emph{et al.}~\cite{shen2024collab_llms} proposed a model collaboration approach where off-the-shelf models learn to choose among themselves which one will handle what portion of all the input tokens. Tokens are generated by a model and then the model-choice decision is represented as a latent variable. In another approach named Mixture of Experts~\cite{xue2024openmoeearlyeffortopen,dai2024deepseekmoeultimateexpertspecialization,jiang2024mixtralexperts,shazeer2017,zhou2022mixture} all experts are trained simultaneously on the same/similar data. These experts are sub-networks of the mixture that can't be used standalone where the feed-forward layers in a transformer is decomposed into modules called \emph{experts}. Similarly Proxy Tuning~\cite{liu2024tuninglanguagemodelsproxy} and \emph{PATHGOOSE}~\cite{muqeeth2024learningroutespecializedexperts} also require same/similar training datasets and/or fine-tuning methods for all experts to bring out the best performance.

Researchers also proposed various prompt-based methods to enhance an LLM's reasoning ability. These methods focus on tuning the instructions detailing context, inputs, or examples. Approaches in this area include few-shot prompting~\cite{brown-fewshot} where in-context learning is enabled by providing demonstrations in the prompt. In chain-of-thought~\cite{wei-cot,kojima2024-zero-shot-cot} paradigm, one activates intermediate reasoning steps for providing a response which promotes a more complex reasoning capability. Meta prompting~\cite{zhang2024metaprompting} tries to improve language model performance by focusing on format and syntax of the problem at hand, and responses are crafted over specific content details. Syntax is used as a guiding template for the expected response or solution. Other methods that aim at improving inference performance include problem decomposition or chaining~\cite{zhou2023leasttomost,khot2023decomposedprompting,hao2023reasoninglanguagemodelplanning}, abstraction~\cite{zheng2024abstraction}, planning and programming~\cite{ding024everythingthoughts,chen2023program,zhou2024solving} or self-verification~\cite{gero2023selfverification}. Similarly, tree-search approaches like Monte-Carlo Tree Search also break the task into smaller sub-tasks and improve the performance by sampling simpler, individual, and intermediate reasoning steps~\cite{yao2023treethoughts,qi2024mutualreasoningslm,zhang2024accessinggpt4levelmathematical} but employs a structured way in doing so. In reflection~\cite{shinn2023reflexion,wang2023dynamicreflection}, further improvement in next response is achieved by reinforcing language agents through linguistic feedback. The linguistic feedback which is essentially a reward score for previous response, also referred to as self-reflection is provided as context to a language model for improving further responses.

In terms of using LLMx in simulation, Diamantopoulos \emph{et al.}~\cite{blockchain-diamat} proposed using LLM in blockchain simulation where they proposed using LLMs for dynamic simulation scenario generation and malicious code modeling. Liu \emph{et al.}~\cite{liu2024llmatransportation} proposed using LLMs in transportation simulation. They proposed leveraging LLMs to represent individual units in the transportation system such as a traveler or driver. Giabbanelli~\cite{gpt-simulation-giabbanelli} and Wu \emph{et al.}~\cite{wu2023smartagentbasedmodelinguse} also proposed smart agent based modeling and simulation methodologies where the LLMs are leveraged in agent-based modeling. 

In our work, we aim to couple off-the-shelf models in a rule-based way to work collaboratively towards common solution by breaking down a problem into simpler sub-tasks. It also inherently enables deployment of hundreds of agents through MPI support. Decomposing the problem also enables exploring new PDES-based algorithms to solve a particular problem using less powerful AI models.

%%%%%%%%%%%%%%%%%%%%%%%%%%%%%%%%%%%%%%%%%%%%%%%%%%%%%%%%%%%%
%%%%%%%%%%%%%%%%%%%%%%%%%%%%%%%%%%%%%%%%%%%%%%%%%%%%%%%%%%%%
%%%%%%%%%%%%%%%%%%%%%%%%%%%%%%%%%%%%%%%%%%%%%%%%%%%%%%%%%%%%

\section{Background}
\label{sec:background}
In this section we discuss some concepts used in the paper.

\subsection{Parallel Discrete Event Simulation}
Parallel discrete event simulation~(PDES) refers to the execution of a single discrete event simulation program on multiple processors~\cite{fujimoto-pdes} so that the simulation runs more efficiently in parallel than traditional sequential simulation. In a discrete event simulation~(DES), a simulated system is decomposed into smaller independent subsystems, and it is assumed that changes in system state occur at discrete points of simulated time. In a popular implementation style, each subsystem would then maintain its own event list with timestamps, containing the events that are scheduled to occur within that subsystem in the future.

The occurrence of events (execution of subsystems) triggers a set of events which may include performing calculations, updating state variables of the simulation and generating new events. This causes the simulation to advance. The simulation runs until all events are processed. In sequential DES, timestamps denote when the event will occur changing the system. The main loop of the simulation engine repeatedly processes the event with smallest timestamp and removes the event from event list~\cite{fujimoto-pdes}. But if an event with larger timestamp (E\textsubscript{t-large}) is executed before a smaller one (E\textsubscript{t-small}), (E\textsubscript{t-large}) might change a state variable used by (E\textsubscript{t-small}) thus making a future event affecting past event. This error is called \emph{causality} error.

PDES tries to execute the events in parallel avoiding causality error (in conservative scheme) or detecting causality error for later recovery (in optimistic scheme)~\cite{fujimoto-pdes, sunil-simx, eker_load-aware_2021-2}. This parallel execution along with clever problem decomposition enables PDES to run very large-scale simulations on large supercomputers~\cite{fujimoto-pdes-network,perumalla-wrap-bg,bauer-warp-bg,barnes-warp,rong-miniSSF}.

\subsection{Language Models}
Language models are a class of generative AI systems designed to understand, synthesize, and manipulate human language~\cite{chang-survey}. They are built using techniques from natural language processing (NLP), a subset of machine learning models and trained on text data to learn the statistical relationship between words or phrases~\cite{gao-stat-lm,conneau-supervised-nlp,devlin-bert, kombrink2011recurrent}. These models are capable of predicting and generating new text sequences based on given input text prompts -- although they face many challenges such as difficulty in understanding complex language, unseen words, structured applications such as mathematics, logic, reasoning etc.

Early language models such as n-gram~\cite{n-gram} (e.g. bigram, trigram) were based on simple statistical modeling techniques in predicting the probabilities of next word based on preceding words. Although these models were capable of question answering, text analysis, summarizing and information retrieval~\cite{barzilay-lee-2004-catching,berger-info-ret,croft2003stat,liu2005statistical}, these models were limited in understanding context from large text.

Later, with the rise of deep learning and GPGPUs, researchers proposed recurrent neural network (RNN) based language models that were capable of capturing temporal relationship among the words in a sentence~\cite{mikolov2010recurrent}. These models were good in tasks like language recognition and translation~\cite{Hori2018speech_rnn,kombrink2011recurrent} but still had problems like vanishing gradient~\cite{bengio-vanishing-grad} and inability to capture very long-range relationships among the words. Researchers adopted long short-term memory (LSTM)~\cite{lstm-rnn} and gated recurrent units (GRUs)~\cite{gru-rnn} to extend RNN based models to address these issues but they were still limited in handling very large sequences of data.

The introduction of the self-attention based transformer architecture~\cite{vaswani-transformer} for language models revolutionized this field by enabling training on very large web-scale unstructured data in parallel and capturing long-rage relationships in the text. Built on transformer architectures, large language models like BERT~\cite{devlin-bert} and GPT~\cite{radford2018gpt} set new standards for solving a wide range of NLP tasks. These models learn broad language representations and can be fine-tuned for specific tasks with smaller, labeled datasets. Because these models are capable of in-context learning, they can generate coherent and contextually relevant responses making them suitable for interactive applications in diverse scenarios~\cite{brown-fewshot}.are symbiotically edusing the Simian PDES semantics at runtime
%%%%%%%%%%%%%%%%%%%%%%%%%%%%%%%%%%%%%%%%%%%%%%%%%%%%%%%%%%%%
%%%%%%%%%%%%%%%%%%%%%%%%%%%%%%%%%%%%%%%%%%%%%%%%%%%%%%%%%%%%
%%%%%%%%%%%%%%%%%%%%%%%%%%%%%%%%%%%%%%%%%%%%%%%%%%%%%%%%%%%%

\begin{figure}[h]
	\centering
    \includegraphics[scale=0.45]{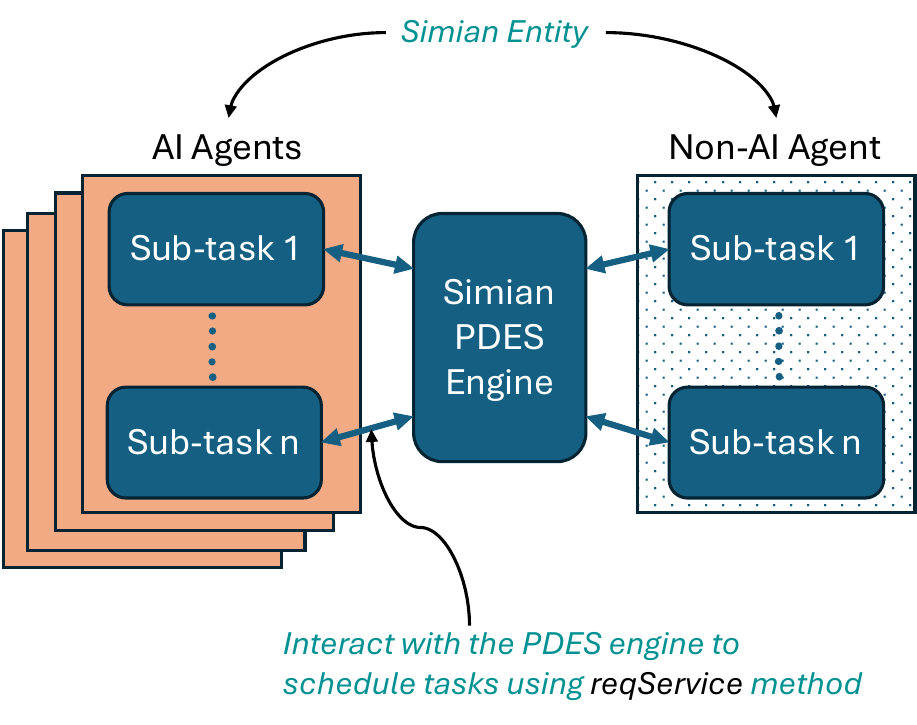}
    \caption{AI / non-AI agents are coupled using the Simian PDES semantics at runtime.}
    \label{fig:method}
\end{figure}

\begin{figure*}[tbp]
\centering
\lstset{style=code_style}
\begin{lstlisting}[language=Python,numbers=left]
# Gathering simulation involving Small Language Model
agent_speed = [10, 15, 20, 25, 30]
init_pos = [[0, 108], [0, 462], [335, 571], [543, 285], [335, 0]]
startTime, endTime, minDelay, useMPI, mpiLib,  simProgress  = 0, 100000, 0.0001, True, "/path/to/libmpich", 1
simianEngine = Simian("Gathering_Agents_Dynamic_Gathering_Positions", startTime, endTime, minDelay, useMPI, mpiLib)
msg_prompt = [ {"role": "system", "content": "You are an AI agent moving in a two-dimensional Euclidean space."},
    {"role": "user", "content": ""},]

class SLMAgent(simianEngine.Entity):
    def __init__(self, baseInfo, *args):
        super(SLMAgent, self).__init__(baseInfo)
        self.position = init_pos[self.num]
        self.max_speed = agent_speed[self.num]
        self.slm = Llama(model_path="Qwen2.5-7B-Instruct-1M-f16.gguf", n_gpu_layers=-1, n_ctx=20000, chat_format="qwen")
    def choose_next_step(self, optimal, *args):
        msg_prompt[1]["content"] = f"You are located at position ({self.position[0]}, {self.position[1]}). Your goal is 
            to go to position ({optimal[0]}, {optimal[1]}). You can move maximum {self.max_speed} units in each step.
            What should be your position in the next step? Verify that the distance between your new position and old 
            position didn't exceed {self.max_speed} units. Strictly follow the following format to provide your answer 
            in integer coordinates in the last line of your response. New_Position:(.., ..)."
        response = self.llm.create_chat_completion(messages = msg_prompt, temperature=0.1)
        lines = response["choices"][0]['message']['content'].split("\n")
        for line in reversed(lines):
            if line.find("New_Position:") != -1:
                new_position = re.findall(r'\(.*?\)',line)[-1]
                break
        self.position[0] = int(eval(re.findall(r'\d+\.\d+|\d+',new_position.split(", ")[0])[-1])) #Parse new pos_x
        self.position[0] = int(eval(re.findall(r'\d+\.\d+|\d+',new_position.split(", ")[1])[-1])) #Parse new pos_y
        reached_optimal = True if math.dist(new_position, optimal) < self.max_speed else False
        #Schedule update_non_slm_copy_of_agent_pos on "Non-AI-Agent" entity.
        self.reqService(simProgress, "update_non_slm_copy_of_agent_pos", (self.num, self.position, reached_optimal), 
        "Non-AI-Agent", 0)
    
class NonAIAgent(simianEngine.Entity):
    def __init__(self, baseInfo, *args):
        super(NonAIAgent, self).__init__(baseInfo)
        self.optimal_pos, self.reached_optimal = None, [False] * agentCount
        self.agent_positions = [([None] * 2) for _ in range(agentCount)],
    def update_non_slm_copy_of_agent_pos(self, data, *args): # data = (agent_num, new_pos, reached_optimal?)
        slm_agent_num, self.reached_optimals[slm_agent_num] = data[0], data[2]
        self.agent_positions[slm_agent_num][0] = data[1][0] # agent[num][0], x coordinater
        self.agent_positions[slm_agent_num][1] = data[1][1] # agent[num][1], y coordinate
        # After updating all agent locations (should happen in same virtual time), schedule next event
        if slm_agent_num == 0: # Only one agent needs to request this
            self.reqService(simProgress,"calculate_optimal_gathering_position",self.agent_positions,"Non-AI-Agent",0)
    def calculate_optimal_gathering_position(self, positions_lst, *args):
        if all(self.reached_optimals): #Checks if all agents reached optimal position
            for i in range(agentCount):
                self.reqService(simProgress, "dump_stats", None, "SLM-Agent", i)
        else: #Calculates geometric median
            self.reached_optimals
            self.optimal_position = CACL_GEOMETRIC_MEDIAN()
            for i in range(agentCount):
                self.reqService(sim_progress, "choose_next_step", self.optimal_position, "SLM-Agent", i)
    ............
simianEngine.addEntity("Non-AI-Agent", NonAIAgent, 0)
for i in range(agentCount):
    simianEngine.addEntity("SLM-Agent", SLMAgent, i)
#Schedule first event at virtual time 0
simianEngine.schedService(0, "calculate_optimal_gathering_position", init_pos[:agentCount], "Non-AI-Agent", 0)

simianEngine.run()
simianEngine.exit()

\end{lstlisting}
% \vspace{-3mm}
\caption{Code snippet from our simulator to solve the gathering problem involving multiple AI agent entities and a non-AI agent entity. Here the non-AI agent mainly performs structured mathematical side-calculations that SLMs are not so good at natively.}
% \vspace{-3mm}
\label{fig:simian_example}
\end{figure*}

%%%%%%%%%%%%%%%%%%%%%%%%%%%%%%%%%%%%%%%%%%%%%%%%%%%%%%%%%%%%
%%%%%%%%%%%%%%%%%%%%%%%%%%%%%%%%%%%%%%%%%%%%%%%%%%%%%%%%%%%%
%%%%%%%%%%%%%%%%%%%%%%%%%%%%%%%%%%%%%%%%%%%%%%%%%%%%%%%%%%%%

\section{Methodology}
\label{sec:methodology}
This section lays out in detail various steps of our proposed approach to couple multiple AI and non-AI agents in a rule-based way. We first explain the coupling technique in subsection~\ref{subsec:coupling_method} followed by how we formulate a problem into the framework of parallel discrete event simulations after applying constraints in subsection~\ref{subsec:problem_formulate}. Finally, we show several case studies from four different domains to demonstrate the effectiveness of the coupling technique in subsection~\ref{subsec:case_studies}.

\subsection{Coupling Technique}
\label{subsec:coupling_method}
We use the open-source Simian~\cite{simian} PDES engine designed for large-scale computing environments to couple multiple AI and non-AI agents. We use the Python implementation of Simian for rapid prototyping of our proposed coupling methodology because it works very well with mainstream LLM libraries such as TensorFlow, PyTorch, llama.cpp, and Transformers. Simian uses entities as objects that contain event handling functions and these entities are automatically distributed among the various MPI ranks for parallel processing. As shown in Figure~\ref{fig:method}, each AI and non-AI agent is implemented as an \textbf{\emph{entity}} of the Simian engine enabling them to inherently run in parallel. The \emph{entity} class has several methods like \emph{attachService, reqService, createProcess, startProcess} to handle the events and processes. In our work we use the \emph{reqService} method to schedule future events at the corresponding \emph{entity} which effectively inserts new events into that \emph{entity's} event list at the appropriate time slots.

Figure~\ref{fig:simian_example} shows an example of how AI/SLM and non-AI agents are defined as Simian \emph{entities}. We present the gathering problem in geometry domain where multiple AI-based agents each with it's own maximum speed try to gather. In the example, a Simian PDES engine is initiated in line 5. Two classes \emph{SLMAgent} and \emph{NonAIAgent} are defined in lines 9 and 35 where both inherit the PDES engine's \emph{Entity} class. One instance of \emph{NonAIAgent} class and multiple instances of \emph{SLMAgent} class are added to the PDES engine as \emph{entities} using the \emph{addEntity} function while also assigning specific names and numbers to those \emph{entities} for identification. In lines 58 to 60 these entities are added. In line 62, the first event of the simulation (\emph{calculate\_optimal\_gathering\_position}) is scheduled at virtual time 0 on the non-AI entity which calculates the \emph{geometric median} of given coordinates. Further events are generated and registered to the corresponding \emph{entities} using the \emph{reqService} method in lines 31, 46, 50, and 55. The simulation starts with \emph{simianEngine.run ()} at line 64 where it starts executing the registered events based on their virtual timestamp. In line 6 and 16 we define the system and user prompts that are passed to language model inference API \emph{create\_chat\_completion} (line 21) along with other parameters such as \emph{sampling temperature}, where the temperature defines the randomness of the output of language models. Higher temperature results in more random and diverse outputs. The language model in the \emph{SLMAgent} is initiated in line 14 where we pass the model path, number of model layers to be offloaded to any available GPU, context size during model inference, and chat format. The response from the language model is parsed and necessary information in extracted which is later passed to further events when registering them using \emph{reqService} method. Simulation ends when no further events are available in the event queue for processing.

Incorporating probabilistic AI models such as the language models inherently adds the non-deterministic characteristics in the simulation. Thus, the simulation scenarios with non-deterministic or somewhat random agent decision requirement, can benefit from the AI/non-AI coupling the most.

\subsection{Problem Formulation}
\label{subsec:problem_formulate}
To solve a problem using SLM involving PDES, we (re)formulate it into a multi-step task by breaking the problem into simpler sub-tasks that can have one of a finite number of possible solutions identifiable by an AI-agent. This way, the AI-agent is constrained to a certain finite solution space, thereby boosting chances of success in each sub-task. When we use more sophisticated models like \emph{Qwen}~\cite{qwen2025}, for simpler tasks the model is often able to provide correct response. In that case we don't constrain the model response by providing it with relevant choices to choose from. We use non-AI agents in tandem with SLMs to solve each sub-task as one of those finite possible choices. When an SLM is asked to solve any sub-task, the non-AI agents may be used to verify and correct the responses of an SLM. On the other hand, complex mathematical sub-tasks that SLMs often fail to provide correct to, can be handled by the non-AI entities. Each sub-task is designated as an \emph{event} of the PDES simulation. Each \emph{event} is associated with a Simian \emph{entity}. When an event is requested, it is added to the corresponding entity's event list along with timestamp information. When the simulation runs with MPI enabled, these entities can opportunistically run methods in parallel and execute independent events with the current timestamps in their event lists.

In many scenarios, it is advantageous to considerably constrain an SLM's solution space -- this helps reduce its tendency to hallucinate or wander-off into irrelevant contexts. As mentioned, one effective strategy we adopted was to provide the SLM with a set of multiple choices to choose from. Furthermore, the many non-AI agents in the simulation are always cross-checking the SLMs' outputs, and working to further dynamically constrain the various AI agent's solution search-space.

As an example in Figure~\ref{fig:multiplication}, the task of multiplication is broken down into the simpler sub-tasks of long multiplication -- single digit multiplication and left-shifts. When a language model is asked to multiply large numbers by itself, it often cannot perform the task correctly, except when the numbers are relatively small~\cite{llm-bad-math, llm-math-ankit}. This behavior is more noticeable for SLMs where the number of parameters is relatively smaller. But we were able to successfully complete the task by breaking down the problem into simpler sub-tasks and involving both SLM and non-AI agents. We perform partial multiplication with all digits of the multiplier in parallel using multiple SLMs deployed as individual entities on multiple nodes. These sub-tasks or events are added to corresponding entities to be executed in chronological (causal) order.

\subsection{Case Studies}
\label{subsec:case_studies}
In this subsection we elaborate how we solve four algorithmic problems from different application domains to demonstrate our coupling methodology. We solve gathering problem from geometry, sorting problem from combinatorics, multiplication problem from arithmetic, and breadth-first traversal problem from graph theory. Even though the chosen problems are relatively simple, and several solution strategies are already available to humans, these are still substantially challenging to solve for most SLMs and even for the prominent mammoth language models. Hence any improvement in the state-of-the-art in enabling algorithmic reasoning tasks using language models can be considered significant. In many ways, one may argue that the chosen problems are representative of some of the harder problems for the current generation of language models to solve on their own, and hence serve as representative tough-case scenarios. Next we describe our approach in each case in more detail.

% In this paper we focus on the geometry problem where multiple AI agents on a 2D space try to gather by continuously broadcasting their updated position and adjusting the optimal gathering position based on their updated position. We show that using non-LLM agents to apply constrains, we can verify LLM outputs and get improved simulation output.

% Later we also show that several other problems can also be solved with the our approach that SLM alone is not very good at solving.

\begin{figure}[b]
	\centering
    \includegraphics[width=\columnwidth]{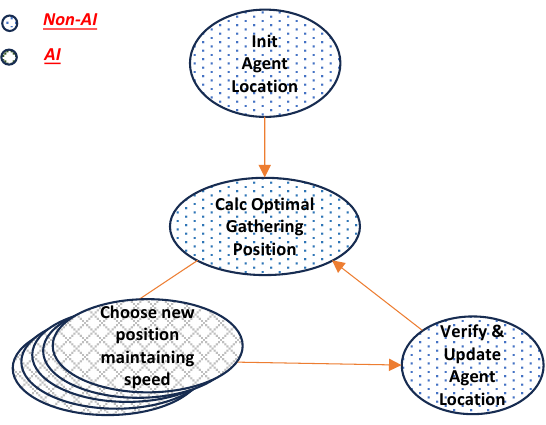}
    \caption{Simulation process of multiple agents trying to gather in a 2D space involving SLM.}
    \label{fig:2d-gahter}
\end{figure}

\subsubsection{Geometry}
We define the problem as \emph{n} number of autonomous AI agents on a 2D space (Euclidean plane) trying to gather starting from their initial positions. Each agent has its own maximum speed, thus limiting the number of units it can move in any direction in a given time-step. The optimal gathering position (geometric median) is calculated and updated by non-AI agent at each time-step. Step by step, our simulation method is as below:

\begin{enumerate}[label=(\alph*)]
\item to start the simulation, the initial positions of the agents are assigned from preset positions so that the same simulation can be performed with varying SLM sampling parameters.
\item the non-AI agent is scheduled to calculate the \emph{geometric median} as optimal gathering position.
\item the SLMs belonging to each agent are asked to choose the new position it should move to in next step following \emph{maximum speed} guideline.
\item the non-AI agent verifies whether the new position provided by the SLM follows \emph{maximum speed} guideline. If not, it sets the correct coordinates.
\item the new positions of the agents are passed to the non-AI agent to further adjust the optimal gathering position.
\end{enumerate}

All of these subtasks are repeated until the agents gather together. We deploy \emph{n}  SLMAgent Simian entities (Figure~\ref{fig:simian_example}, line 9) as \emph{n} gathering agents. These entities run in parallel by deploying multiple instances of SLMs on different MPI ranks. The steps are illustrated in Figure~\ref{fig:2d-gahter} where multiple instances of SLM-based agents run parallelly to determine new agent position.

\begin{figure}[htb]
	\centering
    \includegraphics[width=\columnwidth]{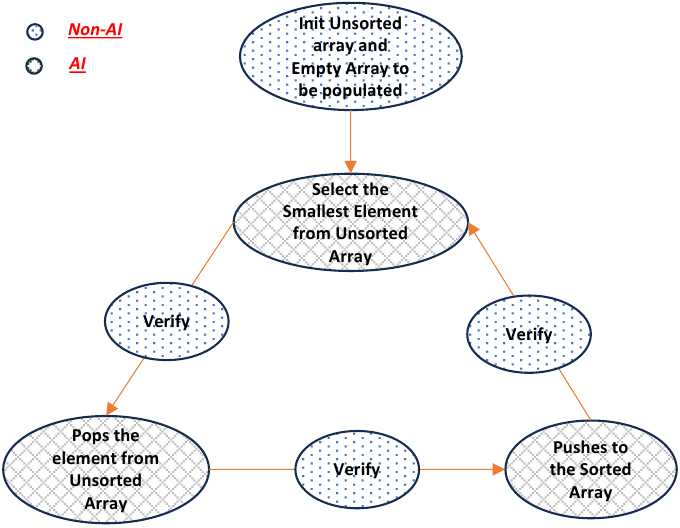}
    \caption{Simulation process for sorting array using SLM.}
    \label{fig:sorting}
\end{figure}

\subsubsection{Combinatorics}
When an SLM is asked to sort an array with broad ranges of numbers, the response is often incorrect. In fact out of 10 times the 8-bit quantized Meta-Llama-3.1-8B-Instruct model was asked to sort a 10-element array, it was able to provide correct answer 6 time. As a workaround, we break down the problem into the following sub-tasks and simulator events mimicking selection sort algorithm to improve accuracy.
\begin{enumerate}[label=(\alph*)]
    \item Non-AI agent initializes the unsorted array and an empty array to hold the sorted array.
    \item SLM based agent is asked to select the smallest element from the unsorted array.
    \item SLM agent removes the element from the unsorted array.
    \item The removed item is appended to the sorted array which is initially empty.
\end{enumerate}

Sub-tasks b, c and, d are repeated until the unsorted array is empty. A non-AI agent is deployed after each of these LLM-based sub-tasks' output to verify those responses and in case of wrong response, it can send feedback to the SLM agent asking for it to fix the erroneous response. One PDES entity is defined as an SLM-based agent, and another as a non-AI agent. These entities execute their corresponding events in parallel which changes the simulation state and adds newly timestamped event requests to that entity's event list. Figure~\ref{fig:sorting} shows a high-level picture of how the problem is formulated as a parallel discrete event simulation.

\subsubsection{Arithmetic}
\begin{figure}[htb]
	\centering
    \includegraphics[width=\columnwidth]{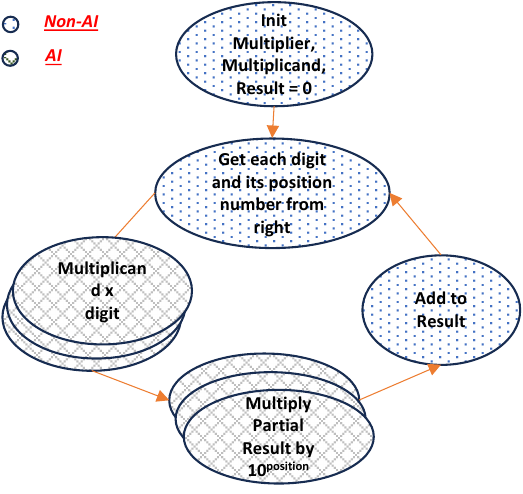}
    \caption{Simulation process for long multiplication involving SLM.}
    \label{fig:multiplication}
\end{figure}

In tandem with LLMs' overall poor performance when solving most structured mathematical problems, SLMs cannot even multiply large numbers correctly~\cite{llm-bad-math,llm-math-ankit}. We break down the task into the simpler steps of long multiplication. Figure~\ref{fig:multiplication} shows our formulation of the problem using a PDES enabled simulation.

\begin{enumerate}[label=(\alph*)]
    \item Define an event/sub-task to get the multiplier digits and their positions starting from rightmost.
    \item Ask the SLM to provide the product of the multiplicand and each digit of multiplier.
    \item Multiply the partial multiplication result with $10^{position}$ using SLM (i.e., left-shifts).
    \item The partials results from each SLM are finally added using a non-AI agent.
\end{enumerate}

Here also, SLM responses in steps b and c are verified and if not correct, the SLM is given yet another chance to correct itself. If there are \emph{i} number of digits in the multiplier then \emph{i} number of SLM based entities may be deployed to perform steps b and c in parallel for each digit. In our experiments, we deployed just one non-LLM agent to handle PDES events that perform steps a, d along with intermediate result verifications.

\subsubsection{Graph theory}
We perform the breadth first traversal of a graph as follows:
\begin{enumerate}[label=(\alph*)]
\item initialize the graphs using a non-AI agent. We use the random graph generator~\cite{nx-random-graph} (\emph{fast\_gnp\_random\_graph}) to randomly generate \emph{10} graphs each with \emph{10} nodes. We use `incident' encoding (using a graph's incidence matrix as opposed to its adjacency matrix) to represent the graphs as text since this encoding has been show to outperform other schemes~\cite{fatemi2024graph}.
\item also initialize an empty queue and an empty list of visited nodes.
\item beginning from the start node, for each of its adjacent nodes, the SLM is asked to check if it is in the \emph{Visited} list. If not, it is pushed to the back of the queue data-structure.
\item next, the queue head is popped and its adjacent nodes are also pushed to the queue depending on the responses from the SLM.
\end{enumerate}

%\atanu{will need to reformulate the simulation to involve SLM more but for now this remains ok.}

\begin{figure}[htb]
	\centering
    \includegraphics[width=0.75\columnwidth]{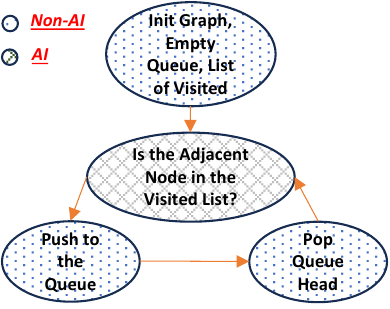}
    \caption{Simulation process for breadth-first traversal involving SLM.}
    \label{fig:bfs}
\end{figure}

\subsection{Open Source Generic PDES-AI Framework}
Our Simian PDES based AI/non-AI hybrid agent simulator has been designed in a generic manner with a view to enable its use in complex, real-world reasoning problems of broader research interest, more diverse than the relatively simple, yet representative case-studies presented in this paper. The authors of this paper intend to open-source the entire PDES-AI simulator framework code for research purposes.

%%%%%%%%%%%%%%%%%%%%%%%%%%%%%%%%%%%%%%%%%%%%%%%%%%%%%%%%%%%%
%%%%%%%%%%%%%%%%%%%%%%%%%%%%%%%%%%%%%%%%%%%%%%%%%%%%%%%%%%%%
%%%%%%%%%%%%%%%%%%%%%%%%%%%%%%%%%%%%%%%%%%%%%%%%%%%%%%%%%%%%
\section{Experiments and Results}
\label{sec:results}
In this section we discuss the experimental setup and results in detail.

\subsection{Hardware Setup}
We use a cluster of six x86 computer nodes connected in a single network to run the MPI enabled parallel discrete event simulations. Each node has an Nvidia Volta GPU~\cite{v100} from the Volta series architecture~\cite{volta-v100} attached to it. Table~\ref{tab:node_config} shows the configuration of each node.

\renewcommand{\arraystretch}{1.5}
\begin{table}[htbp]
    \centering
    \caption{CPU, GPU and Memory Configuration of a compute node.}
\begin{tabular}{|P{1.3in}|c|c|P{0.5in}|}
    \hline
    CPU & RAM & GPU & GPU Memory\\
    \hline
    Intel(R) Xeon(R) CPU E5-2660 v3 @ 2.60GHz & 125GB & Tesla\_V100S & 32GB\\
    \hline
\end{tabular}
\label{tab:node_config}
\end{table}
\renewcommand{\arraystretch}{1}

% TODO: rewrite scaling, experiment still running
To demonstrate the scaling gain from deploying the models in multiple ranks for parallel execution, we use a different hardware setup. This is due to the fact that when we offload the models to the attached GPUs listed in Table~\ref{tab:node_config} to accelerate token generation, GPU memory runs out and thus multiple MPI ranks can't be assigned to single node to show strong scaling. Due to this we run the models on a cluster of 6 AMD EPYC\_7402 @2.8GHz CPUs each with 503GB of memory attached. This larger memory size allows us to assign multiple MPI ranks on a single node avoiding overflow. We take the geometry problem to show simulation scaling with MPI rank increase. We assign 18 CPU threads for the SLM inference within each MPI rank. Thus each MPI rank deploys 18 threads to run inference in CPUs.

\subsection{SLM Models}
We use \emph{instruct} versions of three state-of-the-art open-weight models -- Qwen~\cite{qwen2025}, Llama3~\cite{dubey2024llama3} and Mistral-Nemo~\cite{mistral-nemo} -- which have been fine tuned for a question-answer scenario. We use the smaller versions of these models in our experiment since the larger models of 70B+ parameters require a large amount of local video-ram memory and more powerful GPUs to accelerate that are stipulated to be unavailable. For Qwen and Llama3, we use a  7 and 8 billion parameter versions respectively and for Mistral-Nemo, the available models had a 12B parameter size. The FP32 versions of these models are still large enough to not fit in our GPUs, since we used GPU configurations with 32GB VRAM memories. As a workaround, for Qwen we used 16 bits and, for Llama3 and Mistral-Nemo we used the 8-bit quantized versions of these models in GPT-Generated Unified Format (GGUF)~\cite{gguf} available on Huggingface repositories~\cite{qwen2025,mistral-hf-gguf}. Table~\ref{tab:slm_config} shows the details of these models. Using such quantized, smaller models further enabled running the initial experiments on an Apple M3 MacBook with Metal Shading Language (MSL) acceleration and 64GB total RAM.~\cite{apple-metal}. We use Qwen in the geometry problem, since it can provide new positions of the agents quite accurately compared to other two models. For the other problems we experiment with LLama3 and Mistral-Nemo.

\renewcommand{\arraystretch}{1.5}
\begin{table}[htbp]
    \centering
    \caption{CPU, GPU and Memory Configuration of a compute node.}
\begin{tabular}{|P{1.2in}|P{0.5in}|P{0.5in}|P{0.5in}|}
    \hline
    Name & Parameter Size & Original Model Size & Quantized Model Size\\
    \hline
    Qwen2.5-7B-Instruct-1M-f16.gguf & 7B & 30.5GB & 15.2 GB\\
    \hline
    Meta-Llama-3.1-8B-Instruct-Q8\_0.gguf & 8B & 32.10GB & 8.54 GB\\
    \hline
    \hspace{10pt}
    Mistral-Nemo-Instruct-2407-Q8\_0.gguf & 12B & 49.00GB & 13.02GB\\
    \hline
\end{tabular}
\label{tab:slm_config}
\end{table}
\renewcommand{\arraystretch}{1}

\begin{figure}
\centering
    \begin{minipage}{0.49\columnwidth}
        \subcaptionbox{Temperature 0.1\label{non-ntrvn-t0.1}}
        {\includegraphics[width=\linewidth]{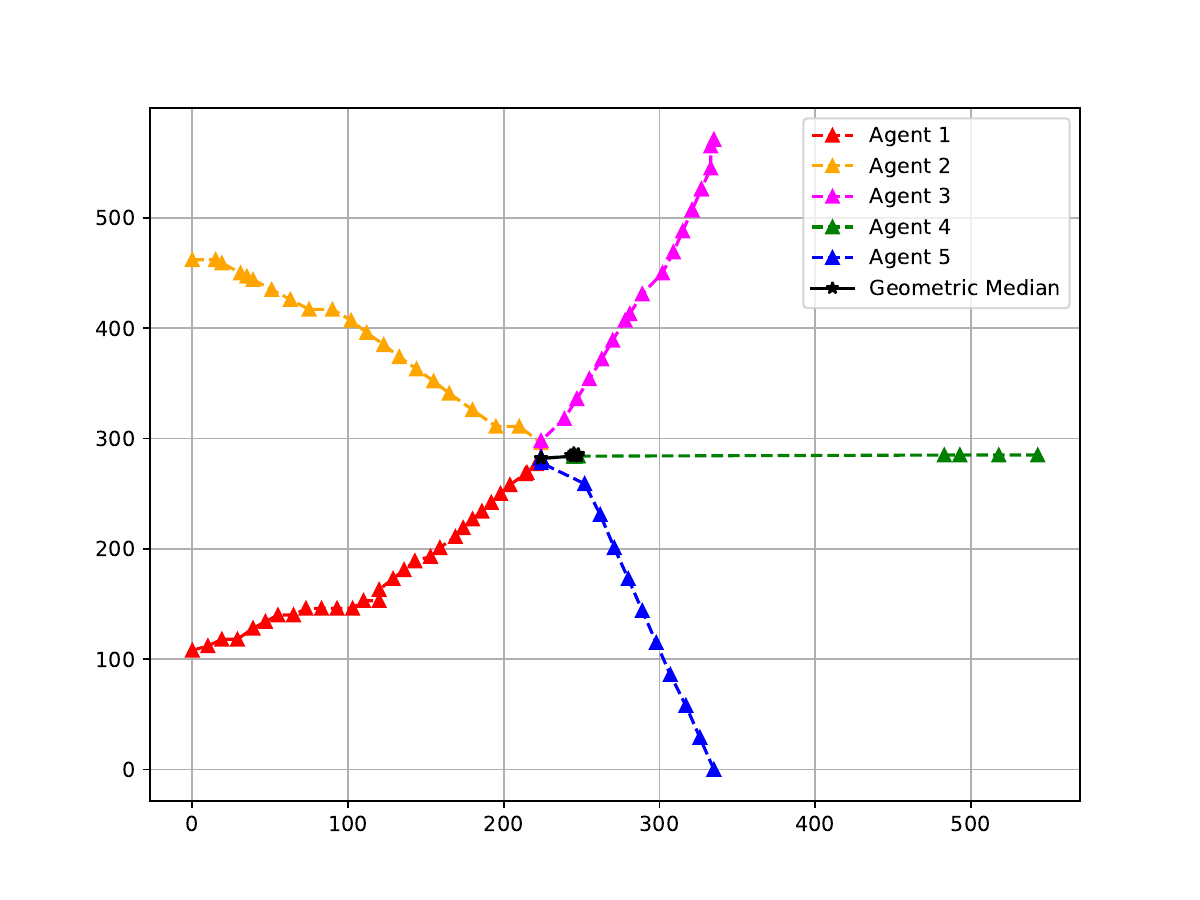}}
        
        \subcaptionbox{Temperature 0.2\label{non-ntrvn-t0.2}}
        {\includegraphics[width=\linewidth]{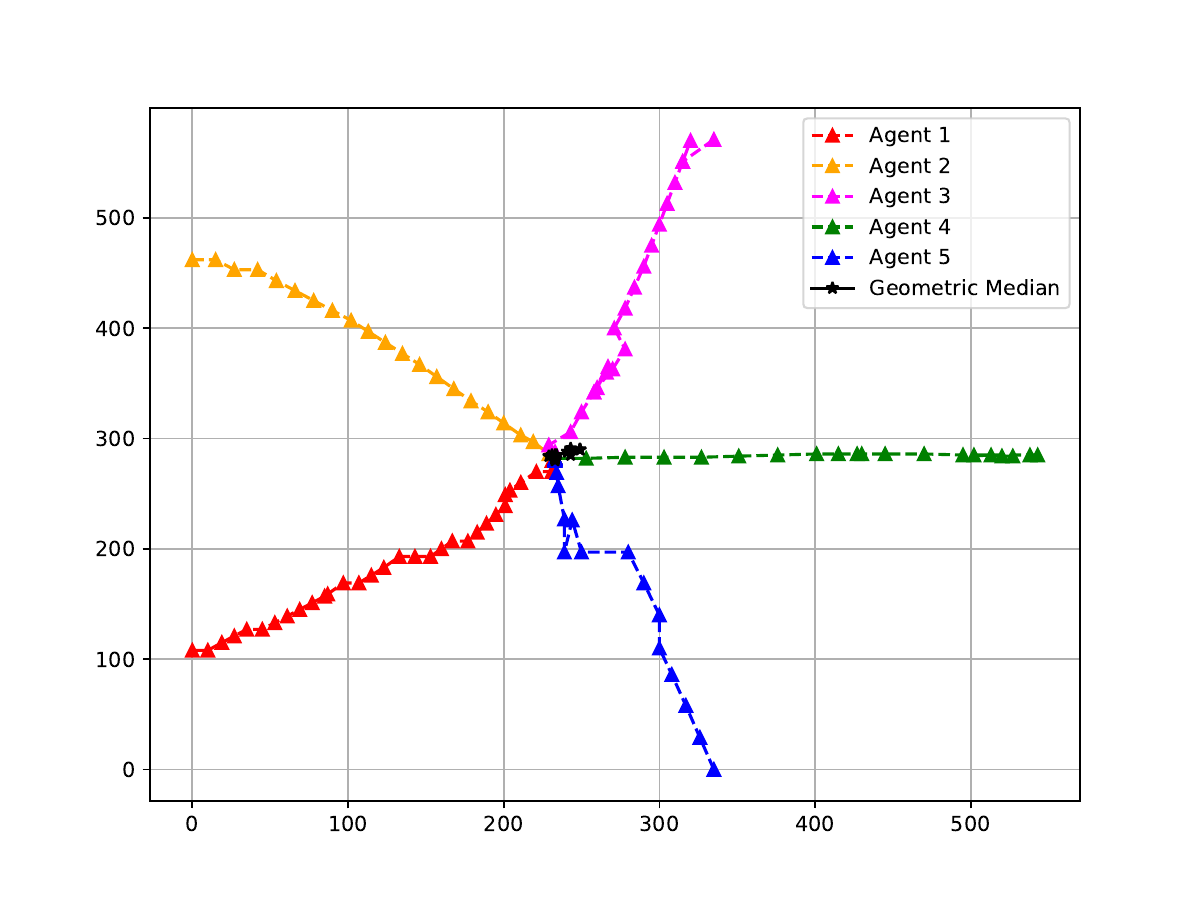}}
        
        \subcaptionbox{Temperature 0.3\label{non-ntrvn-t0.3}}
        {\includegraphics[width=\linewidth]{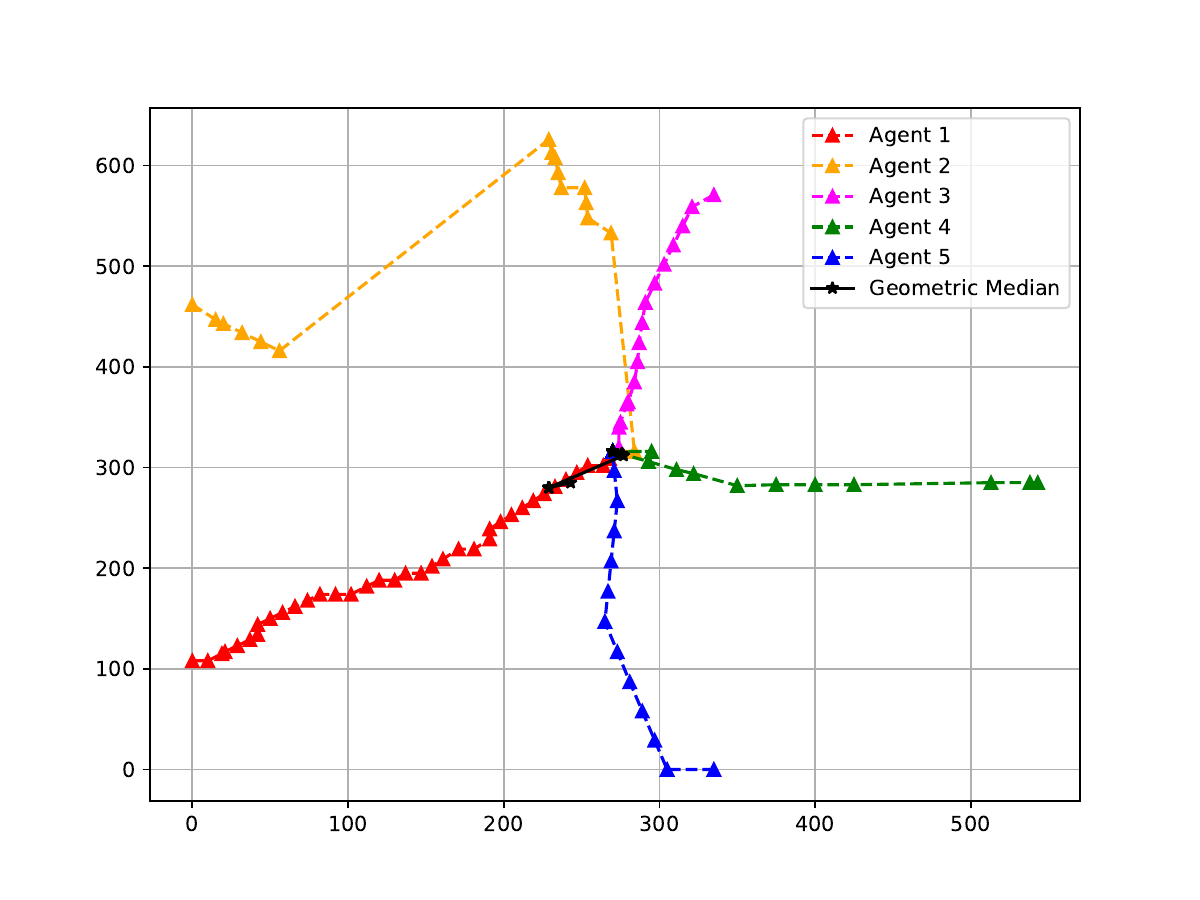}}
        
        \subcaptionbox{Temperature 0.4\label{non-ntrvn-t0.4}}
        {\includegraphics[width=\linewidth]{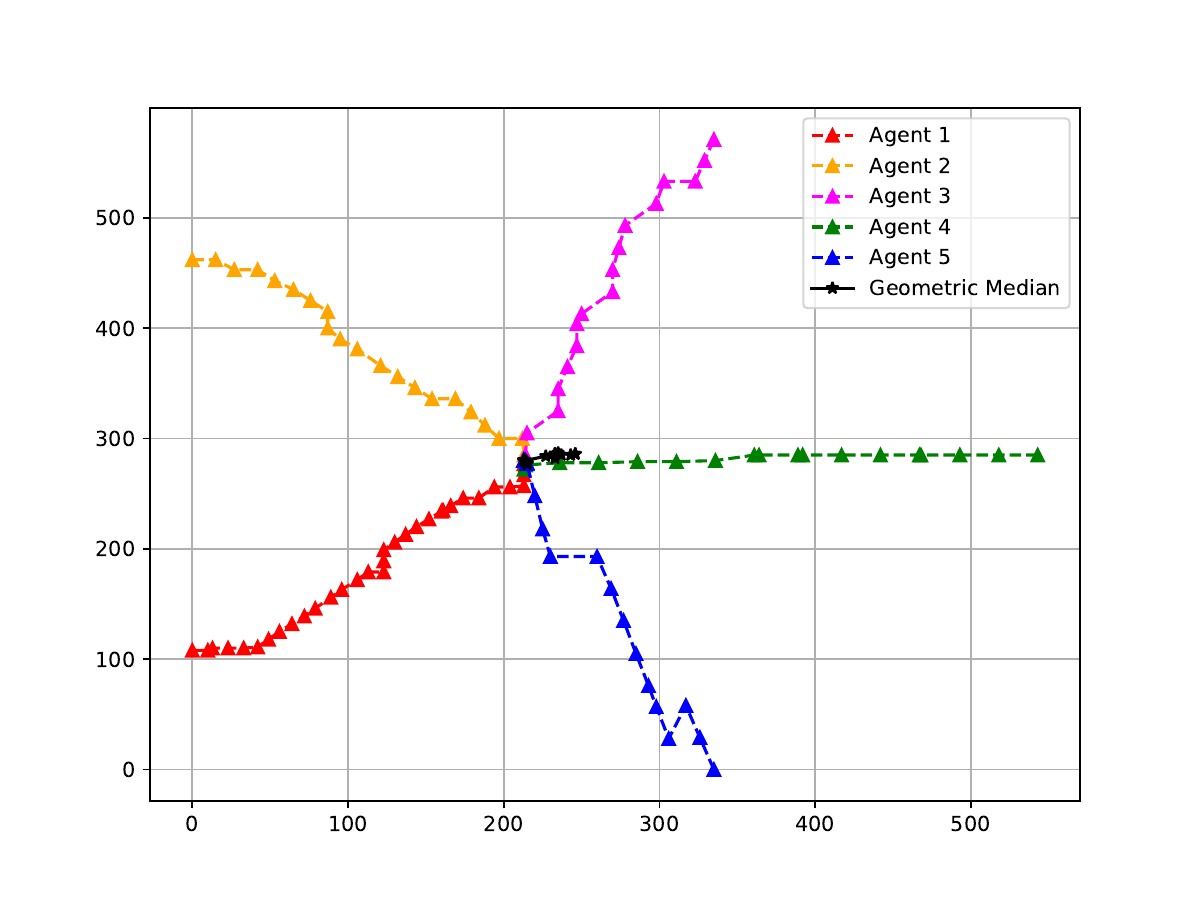}}
        
        \subcaptionbox{Temperature 0.5\label{non-ntrvn-t0.5}}
        {\includegraphics[width=\linewidth]{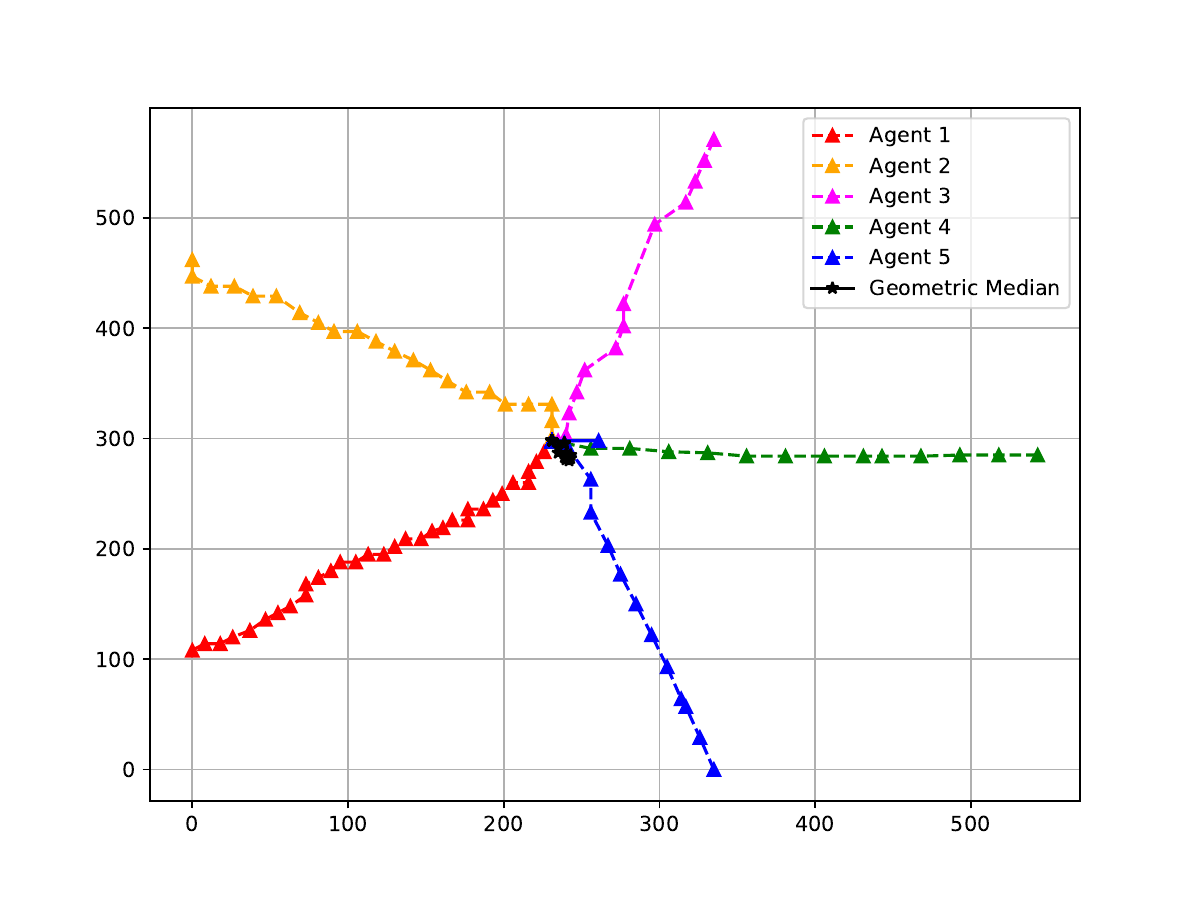}}
    \end{minipage}
    \hfill
    \begin{minipage}{0.49\columnwidth}
        \subcaptionbox{Temperature 0.1\label{ntrvn-t0.1}}
        {\includegraphics[width=\linewidth]{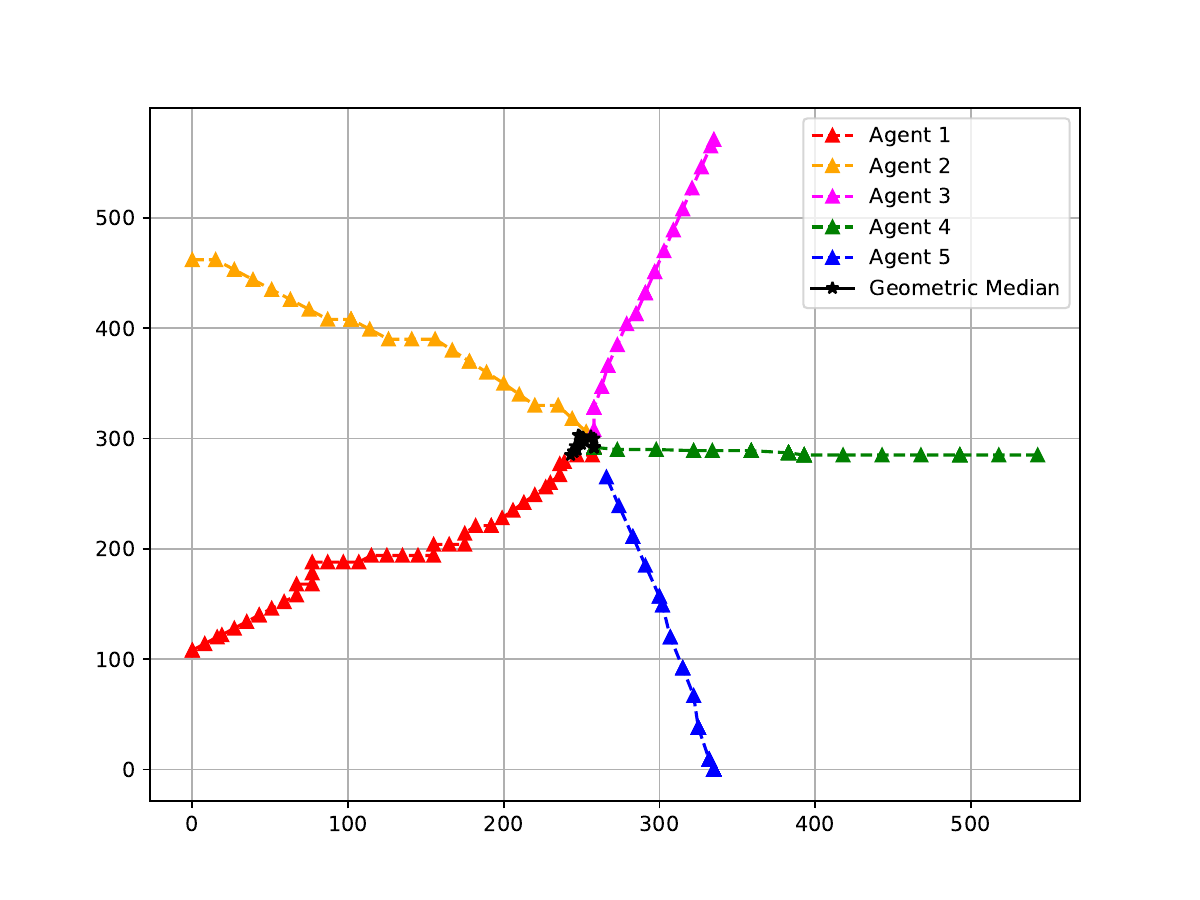}}
        
        \subcaptionbox{Temperature 0.2\label{ntrvn-t0.2}}
        {\includegraphics[width=\linewidth]{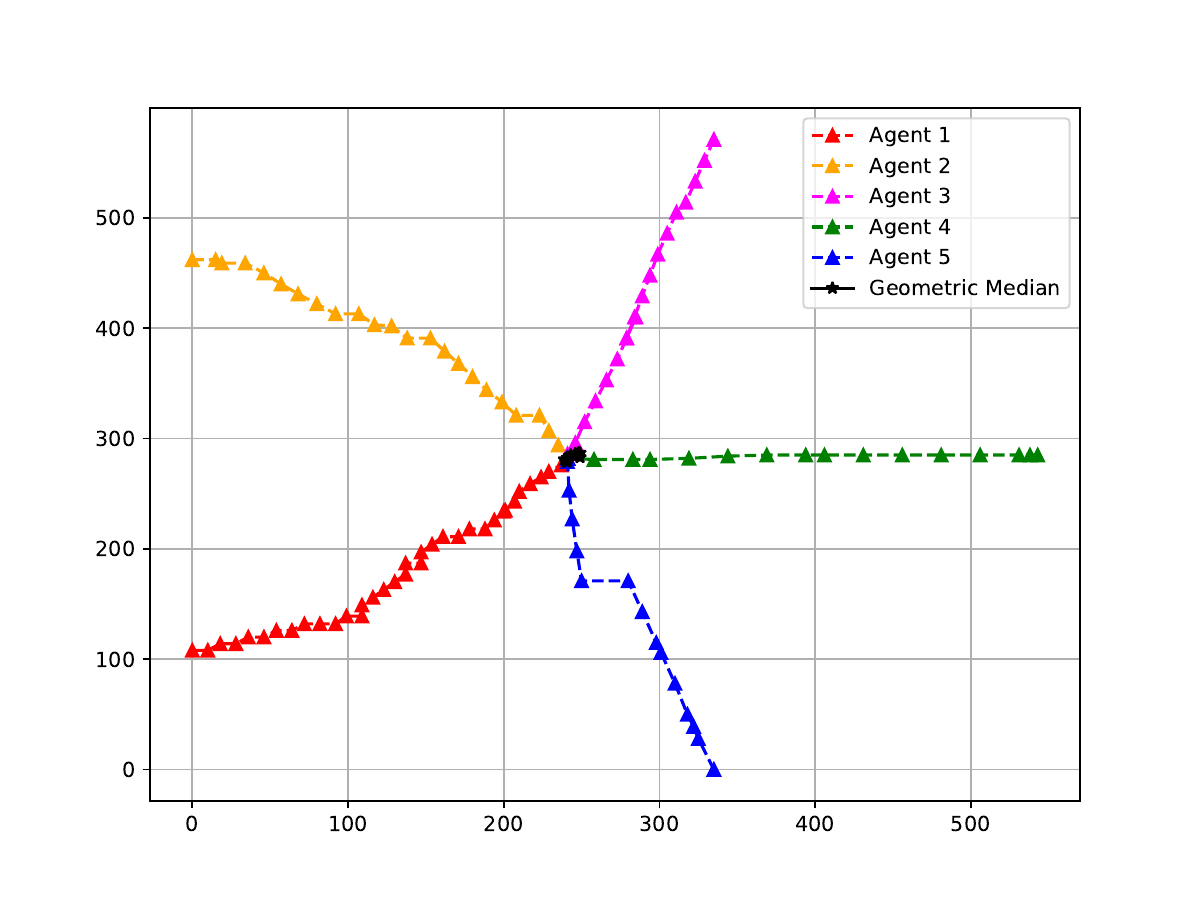}}
    
        \subcaptionbox{Temperature 0.3\label{ntrvn-t0.3}}
        {\includegraphics[width=\linewidth]{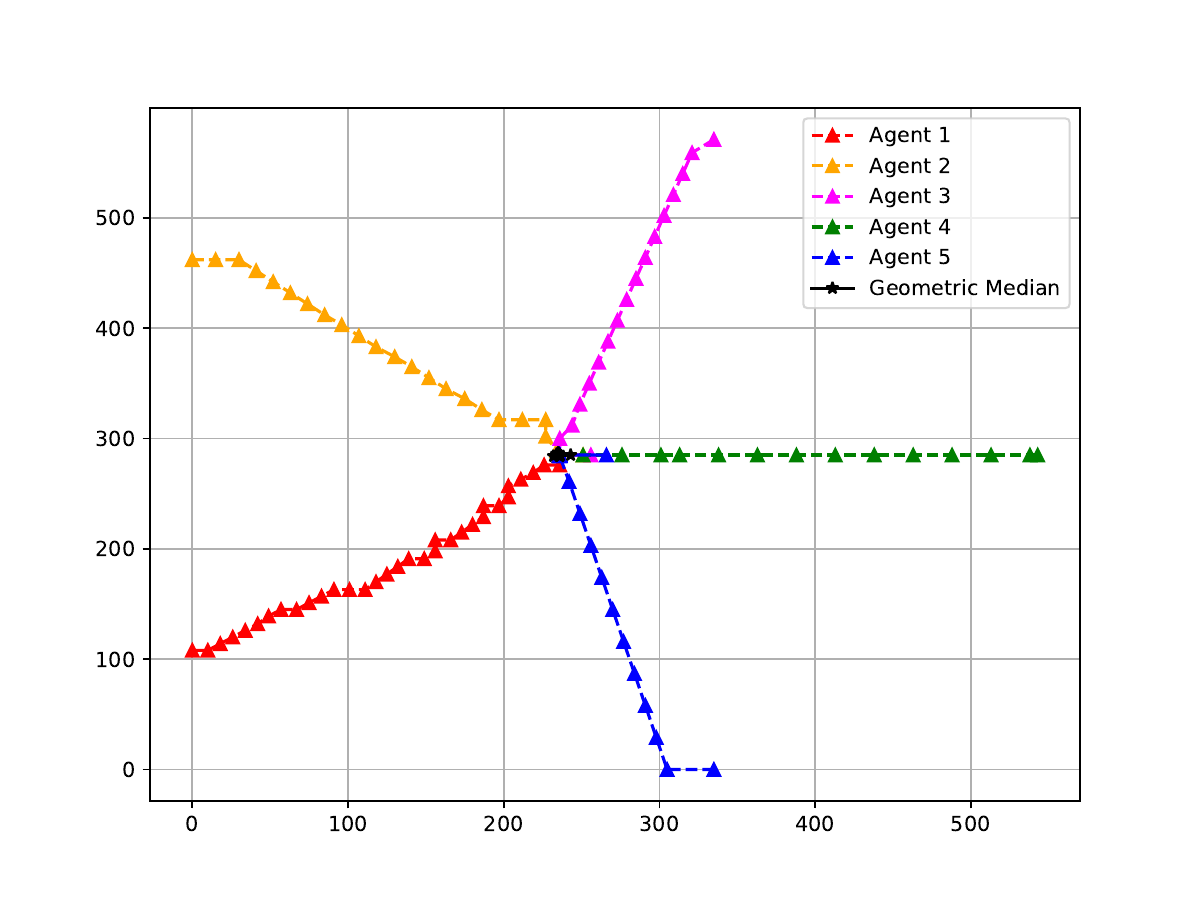}}
    
        \subcaptionbox{Temperature 0.4\label{ntrvn-t0.4}}
        {\includegraphics[width=\linewidth]{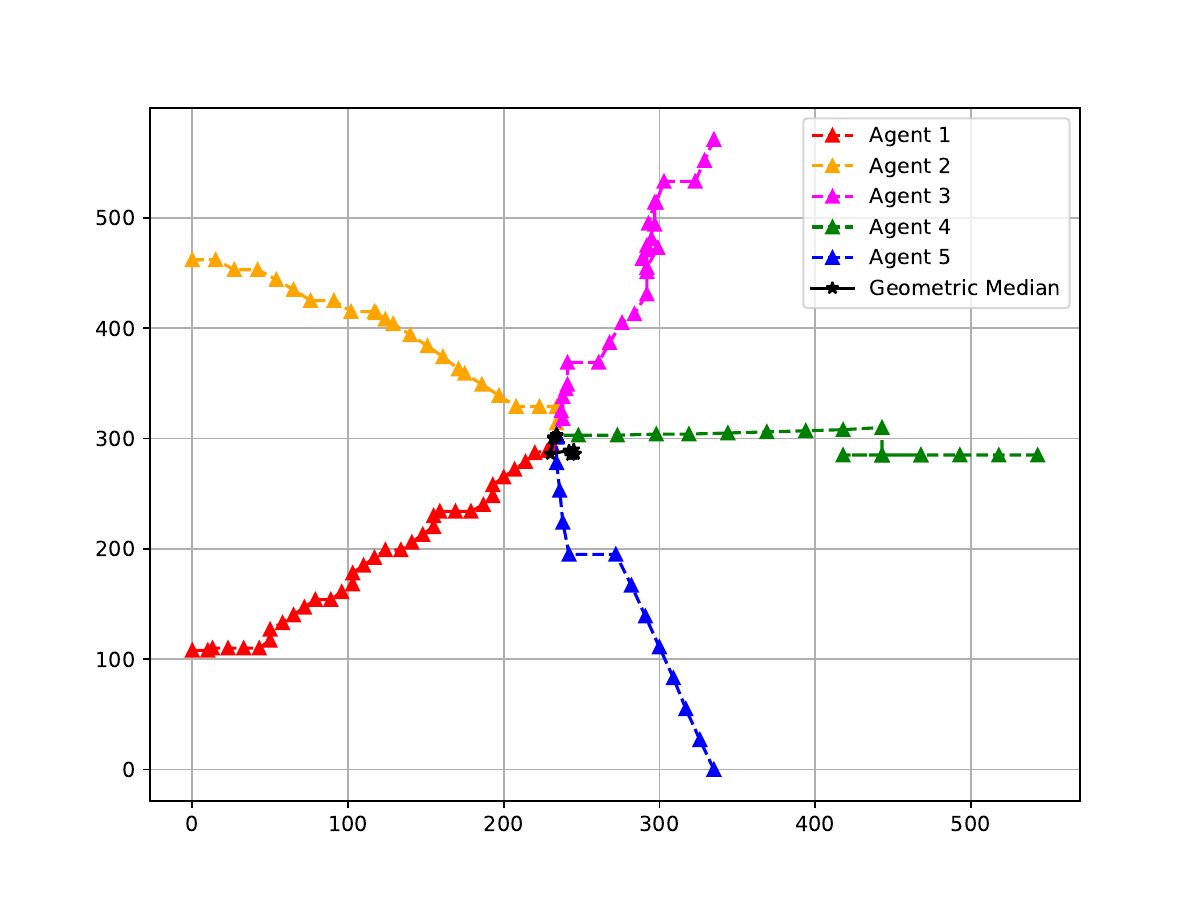}}
    
        \subcaptionbox{Temperature 0.5\label{ntrvn-t0.5}}
        {\includegraphics[width=\linewidth]{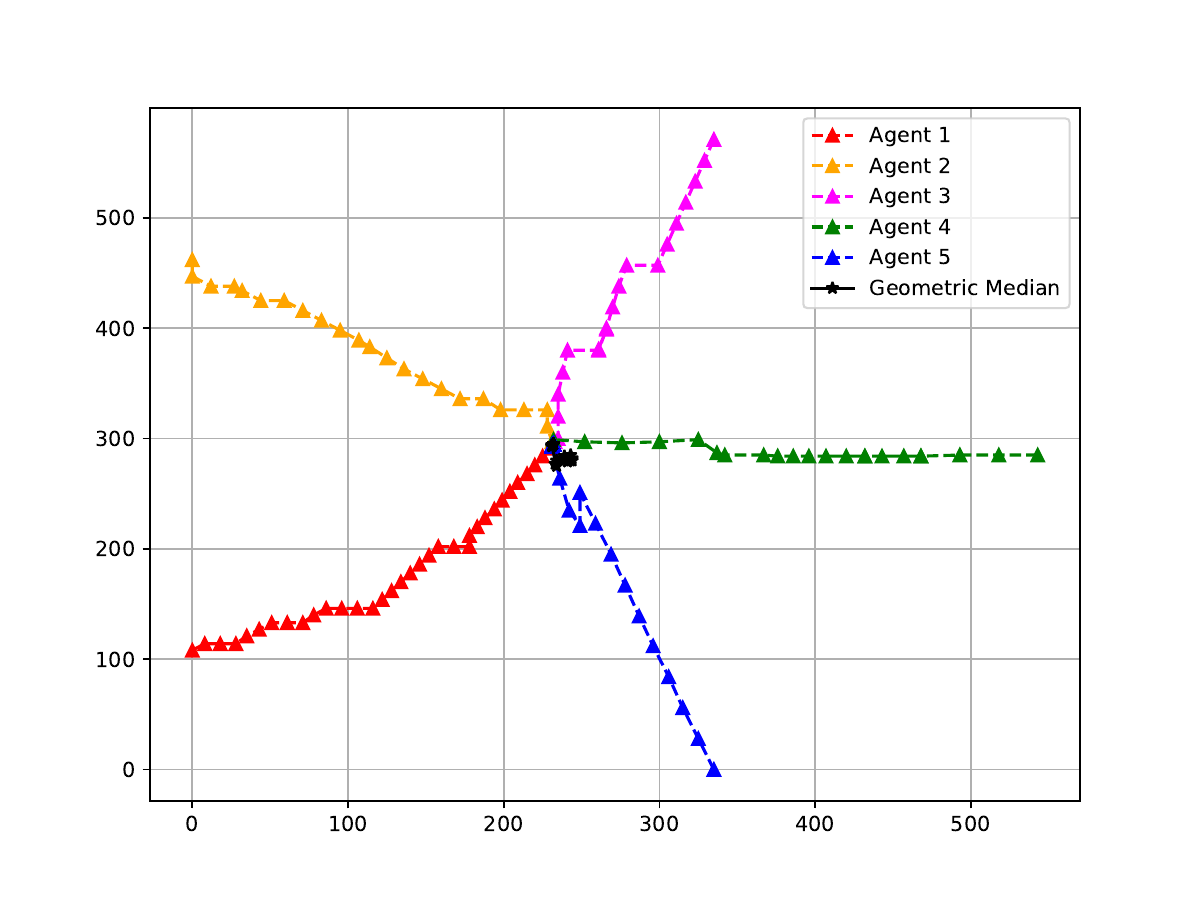}}
    \end{minipage}
\caption{Comparing the paths the SLM-based agents take when the SLM is asked to choose paths varying SLM sampling temperatures. The figures in the left column (a-e) show the path the SLM-based agents take when the non-AI agent / simulation entity doesn't verify and correct the SLM response. The images in right column (f-j) shows the path the same agents take with same SLM temperature settings, but response verified and corrected by the non-AI agent.}
\label{fig:}
\end{figure}

To run inference on these models we used \emph{llama-cpp-python}\cite{llama-cpp-python}, which is a python binding of the \emph{llama-cpp}~\cite{llama-cpp} implementation. Llama-cpp-python is an open-source, lightweight, cross-platform supported, community driven C++ accelerated LLM inference library. It supports quantization and is especially optimized to run efficiently on devices with limited resources. Thus it enables running the models locally even on CPUs without GPU acceleration ensuring on-device privacy where user data never leaves the user premises. When running the SLM inference using \emph{llama-cpp-python}, we keep the default parameters such as \emph{top\_p = 0.95} (sample from tokens with 95\% probability), \emph{top\_k = 40} (sample from top 40 tokens) except for the gathering problem where we vary model\emph{temperature}.

Line 10 in Figure~\ref{fig:simian_example} shows how a model object is initialized using \emph{llama-cpp-python}. The path to the model location and preferred context size is provided to the object-constructor along with the preferred chat format for the model. Llama-cpp-python supports \emph{llama-3} and \emph{mistral-instruct} formats which are required to run the models in table~\ref{tab:slm_config}. Line 21 shows an example how a response can be generated from a language model. In our experimental setup, we use $8192$ as our standard context window size, use GPU acceleration whenever available and further limit the maximum number of tokens generated by the SLM to $512$, leaving other parameters to their defaults.

%%%%%%%%%%%%%%%%%%%%%%%%%%%%%%%%%%%%%%%%%%%%%%%%%%%%%%%%%%%%
%%%%%%%%%%%%%%%%%%%%%%%%%%%%%%%%%%%%%%%%%%%%%%%%%%%%%%%%%%%%
%%%%%%%%%%%%%%%%%%%%%%%%%%%%%%%%%%%%%%%%%%%%%%%%%%%%%%%%%%%%

% \begin{figure*}[htb]
% 	\centering
%     % \includegraphics[width=\columnwidth]{2d-gather.pdf}
%     \includegraphics[width=1.8\columnwidth]{result-comparison.pdf}
%     \caption{Comparison of results among PDES-based approach and vanilla SLM with Zero-Shot and CoT prompt}
%     \label{fig:result}
% \end{figure*}

\begin{figure}[tb]
	\centering
    \includegraphics[width=\columnwidth]{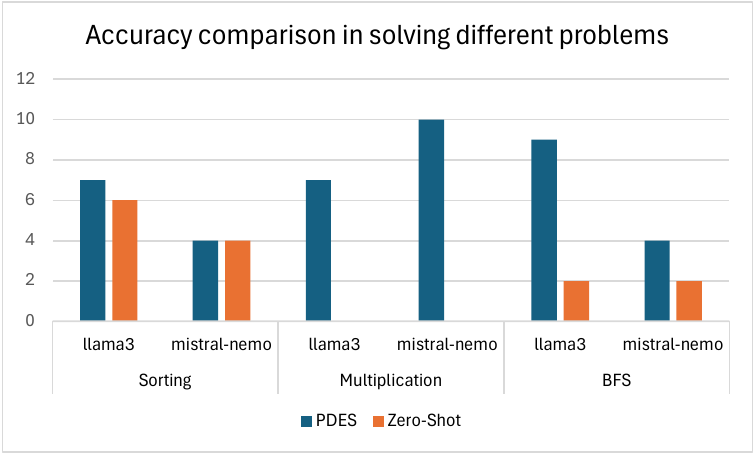}
    \caption{Comparison of results among PDES-based approach and vanilla SLM with Zero-Shot prompt.}
    \label{fig:result}
\end{figure}

\begin{figure}[htb]
% \vspace{-2em}
	\centering
    \includegraphics[width=\columnwidth]{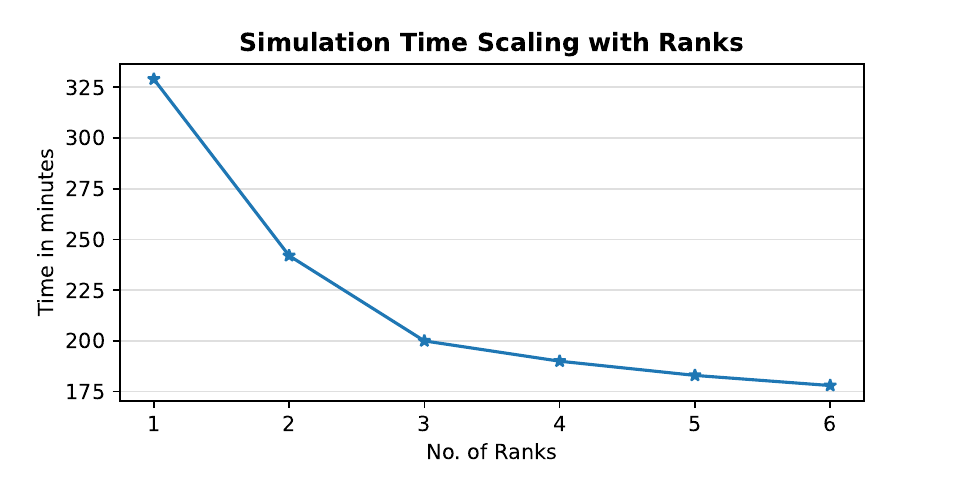}
    \caption{Simulation time as number of MPI ranks in the PDES model is varied.}
    \label{fig:time_scaling}
\vspace{-1.5em}
\end{figure}

\begin{figure}[htb]
% \vspace{-2em}
	\centering
    \includegraphics[width=\columnwidth]{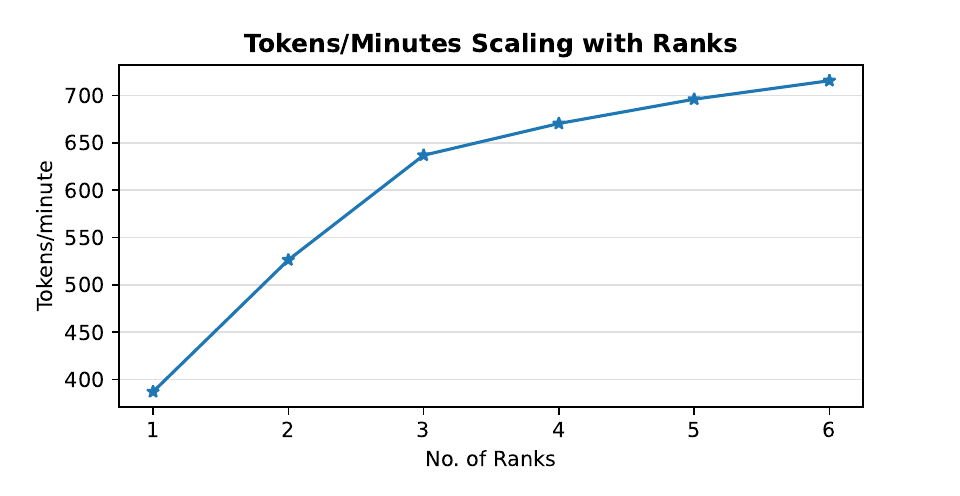}
    \caption{Tokens/sec with scaling of MPI ranks}
    \label{fig:token_scaling}
\vspace{-1.5em}
\end{figure}

\subsection{Result Verification}
To compare the result of our approach against the response solely from the SLM (without PDES based coupling, to serve as the base-case scenario for comparison), we ask the SLM to solve the same problem using zero-shot prompting~\cite{wei2022finetuned-zero-shot}. Here the SLMs are also given a second chance to correct their responses. To serve as a benchmark to compare our coupling methodology against vanilla stand-alone language-models, we randomly generated ten problems for the \emph{combinatorics, arithmetic, and graph} domains mentioned in subsection~\ref{subsec:case_studies} to be solved. As an example, we used the following user and system prompts to get the responses from SLMs for the sorting problem.

\begin{lstlisting}[breaklines=true]
System Prompt: You are an AI agent who can sort an array.
User Prompt: Sort the array [475623, 577963, 156161, 563672, 7440, 844580, 365172, 730079, 984081, 120146].
\end{lstlisting}

% For combinatorics(sorting) problem, the randomly generated 10 arrays are following:
% [475623, 577963, 156161, 563672, 7440, 844580, 365172, 730079, 984081, 120146], 
% [497845, 7731, 126449, 990865, 876935, 108620, 442503, 607567, 880934, 404025],
% [999037, 607114, 450832, 144406, 876971, 906688, 509560, 631147, 16391, 13190], 
% [593417, 43914, 346524, 795660, 465130, 30644, 524785, 747968, 888748, 758244],
% [957058, 99858, 783249, 699602, 697972, 681497, 178098, 343718, 623196, 945340],
% [488798, 778753, 615156, 54068, 863824, 333114, 935701, 379599, 826133, 815198],
% [244271, 365810, 615036, 851601, 242565, 881741, 334407, 818557, 719940, 713915],

For the \emph{Geometry} problem, Figure~\ref{fig:} shows the path of the agents in various conditions. We choose the starting positions of the agents as the vertices of a pentagon for better visualization. These vertices are listed on Figure~\ref{fig:simian_example}'s line 3. The maximum distance each agent can travel is also listed in line 2 of the same figure. The left column of the figure shows the path the agents take when trying to gather. We vary the \emph{sampling temperature} of the LLM agents to obtain different paths the agents take to gather. Without verification of SLM response, we notice that sometimes the agents take big jumps exceeding the predefined \emph{maximum distance} it can travel in each step. On the other hand, the figures in the right column show that when these jumps are detected by the non-AI agent, it can ask the SLM to provide correct \emph{new position} or even it can itself correct the error; thus avoiding the jumps. Overall smoothness and quality of the agent trajectories is observed to be better with the symbiotic coupling strategy, with differences being more noticeable at lower temperatures.

Figure~\ref{fig:result} compares the accuracy of our approach against zero-shot prompting in solving the problems except the geometry(gathering problem). For the array sorting problem we observe that the LLM makes mistake in the \emph{smallest number selection} stage. We see the largest gain in accuracy for the \emph{multiplication} task. We also observe that \emph{llama-3.1} models performs better that \emph{mistral-nemo} in sorting and bfs. In case of multiplication, the accuracy depends on the capability to do single-digit multiplication of the multiplicand. Since the parameter size of \emph{mistral-nemo} is larger than \emph{llama3}, this behavior is expected. Overall, when llama-3.1 SLM is used, PDES based solution has overall accuracy of 77\% where the accuracy of zero-shot prompt is 26\%. On the other hand, for mistral-nemo, the accuracy of PDES based and zero-shot prompt based solution is 60\% and, 20\% respectively.

Figure~\ref{fig:time_scaling} shows that with the increase of simulation ranks, the simulation time decreases while Figure~\ref{fig:token_scaling} shows that the average number of tokens generated by the language models per minute also increase with the increase of MPI ranks for the gathering simulation. The incremental improvements are more when the number of MPI ranks are low, since more agent \emph{entities} need to be assigned on the same MPI rank (or compute node). In the \emph{geometric} simulation scenario which is shown in the plot, there were $6$ agents in total: $5$ SLM AI agents, and $1$ non-AI rule enforcing agent. In all our four scenario based experiments, strong scaling in simulation time and tokens/minute has been observed up to as many ranks as there are instantiated agents, since the AI agents in particular are almost always compute/memory-resource bound.

%%%%%%%%%%%%%%%%%%%%%%%%%%%%%%%%%%%%%%%%%%%%%%%%%%%%%%%%%%%%
%%%%%%%%%%%%%%%%%%%%%%%%%%%%%%%%%%%%%%%%%%%%%%%%%%%%%%%%%%%%
%%%%%%%%%%%%%%%%%%%%%%%%%%%%%%%%%%%%%%%%%%%%%%%%%%%%%%%%%%%%
\section{Conclusion}
\label{sec:conclusion}
This paper presents a PDES based novel methodology to couple multiple AI and non-AI agents in a rule-based way to work towards a common goal. With the support of MPI, careful problem decomposition, and multiple-choice constraints, the AI and non-AI agents run in parallel to solve each sub-task of a more complex parent-task. We demonstrated how four problems from diverse domains may be efficiently solved by our structured, parallel discrete event simulation strategy involving multiple AI and non-AI agents. Moreover, the intermediate results are cross-checked using non-AI agents (potentially involving even more complex sub-simulations), promoting verifiability of AI model responses and thereby ensuring trustworthiness of AI agent simulations. Results show that our approach achieves a much better accuracy while enabling collaborative problem solving, among multiple AI and non-AI agents, which is essentially the first step towards automated harnessing of domain expertise from different fields. Furthermore, we intend to open-source release the PDES-AI agent simulator codebase for research purposes.

% \begin{acks}
% \end{acks}


\bibliographystyle{ACM-Reference-Format}
\bibliography{main-sigconf}


\begin{thebibliography}{81}

%%% the Web address.

\ifx \showCODEN    \undefined \def \showCODEN     #1{\unskip}     \fi
\ifx \showISBNx    \undefined \def \showISBNx     #1{\unskip}     \fi
\ifx \showISBNxiii \undefined \def \showISBNxiii  #1{\unskip}     \fi
\ifx \showISSN     \undefined \def \showISSN      #1{\unskip}     \fi
\ifx \showLCCN     \undefined \def \showLCCN      #1{\unskip}     \fi
\ifx \shownote     \undefined \def \shownote      #1{#1}          \fi
\ifx \showarticletitle \undefined \def \showarticletitle #1{#1}   \fi
\ifx \showURL      \undefined \def \showURL       {\relax}        \fi
% invisible to TeX
\providecommand\bibfield[2]{#2}
\providecommand\bibinfo[2]{#2}
\providecommand\natexlab[1]{#1}
\providecommand\showeprint[2][]{arXiv:#2}

\bibitem[Alonso et~al\mbox{.}(2001)]%
        {alonso2002multiagentlearning}
\bibfield{author}{\bibinfo{person}{Eduardo Alonso}, \bibinfo{person}{Mark
  D’Inverno}, \bibinfo{person}{Daniel Kudenko}, \bibinfo{person}{Michael
  Luck}, {and} \bibinfo{person}{Jason Noble}.} \bibinfo{year}{2001}\natexlab{}.
\newblock \showarticletitle{Learning in multi-agent systems}.
\newblock \bibinfo{journal}{\emph{The Knowledge Engineering Review}}
  \bibinfo{volume}{16}, \bibinfo{number}{3} (\bibinfo{year}{2001}),
  \bibinfo{pages}{277–284}.
\newblock
\href{https://doi.org/10.1017/S0269888901000170}{doi:\nolinkurl{10.1017/S0269888901000170}}


\bibitem[Apple(2024)]%
        {apple-metal}
\bibfield{author}{\bibinfo{person}{Apple}.} \bibinfo{year}{July
  2024}\natexlab{}.
\newblock \bibinfo{booktitle}{\emph{Metal Overview}}.
\newblock
\urldef\tempurl%
\url{https://developer.apple.com/metal/}
\showURL{%
\tempurl}


\bibitem[Baratta et~al\mbox{.}(2023)]%
        {BARATTA2023surv-human-robot}
\bibfield{author}{\bibinfo{person}{Alessio Baratta}, \bibinfo{person}{Antonio
  Cimino}, \bibinfo{person}{Maria~Grazia Gnoni}, {and}
  \bibinfo{person}{Francesco Longo}.} \bibinfo{year}{2023}\natexlab{}.
\newblock \showarticletitle{Human Robot Collaboration in Industry 4.0: a
  literature review}.
\newblock \bibinfo{journal}{\emph{Procedia Computer Science}}
  \bibinfo{volume}{217} (\bibinfo{year}{2023}), \bibinfo{pages}{1887--1895}.
\newblock
\showISSN{1877-0509}
\href{https://doi.org/10.1016/j.procs.2022.12.389}{doi:\nolinkurl{10.1016/j.procs.2022.12.389}}
\newblock
\shownote{4th International Conference on Industry 4.0 and Smart
  Manufacturing}.


\bibitem[Barnes et~al\mbox{.}(2013)]%
        {barnes-warp}
\bibfield{author}{\bibinfo{person}{Peter~D. Barnes},
  \bibinfo{person}{Christopher~D. Carothers}, \bibinfo{person}{David~R.
  Jefferson}, {and} \bibinfo{person}{Justin~M. LaPre}.}
  \bibinfo{year}{2013}\natexlab{}.
\newblock \showarticletitle{Warp speed: executing time warp on 1,966,080
  cores}. In \bibinfo{booktitle}{\emph{Proceedings of the 1st ACM SIGSIM
  Conference on Principles of Advanced Discrete Simulation}} (Montr©al,
  Qu\'{e}bec, Canada) \emph{(\bibinfo{series}{SIGSIM PADS '13})}.
  \bibinfo{publisher}{Association for Computing Machinery},
  \bibinfo{address}{New York, NY, USA}, \bibinfo{pages}{327–336}.
\newblock
\showISBNx{9781450319201}
\href{https://doi.org/10.1145/2486092.2486134}{doi:\nolinkurl{10.1145/2486092.2486134}}


\bibitem[Bartowski(2024)]%
        {mistral-hf-gguf}
\bibfield{author}{\bibinfo{person}{A Bartowski}.} \bibinfo{year}{Septerber
  2024}\natexlab{}.
\newblock \bibinfo{booktitle}{\emph{Mistral-Nemo-Instruct-2407-GGUF - Hugging
  Face}}.
\newblock
\urldef\tempurl%
\url{https://huggingface.co/bartowski/Mistral-Nemo-Instruct-2407-GGUF}
\showURL{%
\tempurl}


\bibitem[Barzilay and Lee(2004)]%
        {barzilay-lee-2004-catching}
\bibfield{author}{\bibinfo{person}{Regina Barzilay} {and}
  \bibinfo{person}{Lillian Lee}.} \bibinfo{year}{2004}\natexlab{}.
\newblock \showarticletitle{Catching the Drift: Probabilistic Content Models,
  with Applications to Generation and Summarization}. In
  \bibinfo{booktitle}{\emph{Proceedings of the Human Language Technology
  Conference of the North {A}merican Chapter of the Association for
  Computational Linguistics: {HLT}-{NAACL} 2004}}.
  \bibinfo{publisher}{Association for Computational Linguistics},
  \bibinfo{address}{Boston, Massachusetts, USA}, \bibinfo{pages}{113--120}.
\newblock
\urldef\tempurl%
\url{https://aclanthology.org/N04-1015}
\showURL{%
\tempurl}


\bibitem[Batagelj and Brandes(2005)]%
        {nx-random-graph}
\bibfield{author}{\bibinfo{person}{Vladimir Batagelj} {and}
  \bibinfo{person}{Ulrik Brandes}.} \bibinfo{year}{2005}\natexlab{}.
\newblock \showarticletitle{Efficient generation of large random networks}.
\newblock \bibinfo{journal}{\emph{Phys. Rev. E}}  \bibinfo{volume}{71}
  (\bibinfo{date}{Mar} \bibinfo{year}{2005}), \bibinfo{pages}{036113}.
\newblock
Issue 3.
\href{https://doi.org/10.1103/PhysRevE.71.036113}{doi:\nolinkurl{10.1103/PhysRevE.71.036113}}


\bibitem[Bauer~Jr. et~al\mbox{.}(2009)]%
        {bauer-warp-bg}
\bibfield{author}{\bibinfo{person}{David~W. Bauer~Jr.},
  \bibinfo{person}{Christopher~D. Carothers}, {and} \bibinfo{person}{Akintayo
  Holder}.} \bibinfo{year}{2009}\natexlab{}.
\newblock \showarticletitle{Scalable Time Warp on Blue Gene Supercomputers}. In
  \bibinfo{booktitle}{\emph{2009 ACM/IEEE/SCS 23rd Workshop on Principles of
  Advanced and Distributed Simulation}}. \bibinfo{pages}{35--44}.
\newblock
\href{https://doi.org/10.1109/PADS.2009.21}{doi:\nolinkurl{10.1109/PADS.2009.21}}


\bibitem[Bengio et~al\mbox{.}(1994)]%
        {bengio-vanishing-grad}
\bibfield{author}{\bibinfo{person}{Y. Bengio}, \bibinfo{person}{P. Simard},
  {and} \bibinfo{person}{P. Frasconi}.} \bibinfo{year}{1994}\natexlab{}.
\newblock \showarticletitle{Learning long-term dependencies with gradient
  descent is difficult}.
\newblock \bibinfo{journal}{\emph{IEEE Transactions on Neural Networks}}
  \bibinfo{volume}{5}, \bibinfo{number}{2} (\bibinfo{year}{1994}),
  \bibinfo{pages}{157--166}.
\newblock
\href{https://doi.org/10.1109/72.279181}{doi:\nolinkurl{10.1109/72.279181}}


\bibitem[Berger and Lafferty(2001)]%
        {berger-info-ret}
\bibfield{author}{\bibinfo{person}{Adam Berger} {and} \bibinfo{person}{John
  Lafferty}.} \bibinfo{year}{2001}\natexlab{}.
\newblock \emph{\bibinfo{title}{Statistical machine learning for information
  retrieval}}.
\newblock \bibinfo{thesistype}{Ph.\,D. Dissertation}. \bibinfo{school}{Carnegie
  Mellon University}, \bibinfo{address}{USA}.
\newblock
\showISBNx{0542045818}
\newblock
\shownote{AAI3168516}.


\bibitem[Betlen(2024)]%
        {llama-cpp-python}
\bibfield{author}{\bibinfo{person}{Andrei Betlen}.} \bibinfo{year}{October
  2024}\natexlab{}.
\newblock \bibinfo{booktitle}{\emph{Python bindings for llama.cpp}}.
\newblock
\urldef\tempurl%
\url{https://github.com/abetlen/llama-cpp-python}
\showURL{%
\tempurl}


\bibitem[Bommasani et~al\mbox{.}(2021)]%
        {Bommasani2021FoundationModels}
\bibfield{author}{\bibinfo{person}{Rishi Bommasani}, \bibinfo{person}{Drew~A.
  Hudson}, \bibinfo{person}{Ehsan Adeli}, \bibinfo{person}{Russ Altman},
  \bibinfo{person}{Simran Arora}, \bibinfo{person}{Sydney von Arx}, {and}
  \bibinfo{person}{Michael S.~Bernstein et. al.}}
  \bibinfo{year}{2021}\natexlab{}.
\newblock \showarticletitle{On the Opportunities and Risks of Foundation
  Models}.
\newblock \bibinfo{journal}{\emph{ArXiv}} (\bibinfo{year}{2021}).
\newblock
\urldef\tempurl%
\url{https://crfm.stanford.edu/assets/report.pdf}
\showURL{%
\tempurl}


\bibitem[Brown et~al\mbox{.}(1992)]%
        {n-gram}
\bibfield{author}{\bibinfo{person}{Peter~F. Brown}, \bibinfo{person}{Peter~V.
  deSouza}, \bibinfo{person}{Robert~L. Mercer}, \bibinfo{person}{Vincent
  J.~Della Pietra}, {and} \bibinfo{person}{Jenifer~C. Lai}.}
  \bibinfo{year}{1992}\natexlab{}.
\newblock \showarticletitle{Class-based n-gram models of natural language}.
\newblock \bibinfo{journal}{\emph{Comput. Linguist.}} \bibinfo{volume}{18},
  \bibinfo{number}{4} (\bibinfo{date}{Dec.} \bibinfo{year}{1992}),
  \bibinfo{pages}{467–479}.
\newblock
\showISSN{0891-2017}


\bibitem[Brown et~al\mbox{.}(2020)]%
        {brown-fewshot}
\bibfield{author}{\bibinfo{person}{Tom Brown}, \bibinfo{person}{Benjamin Mann},
  \bibinfo{person}{Nick Ryder}, \bibinfo{person}{Melanie Subbiah},
  \bibinfo{person}{Jared~D Kaplan}, \bibinfo{person}{Prafulla Dhariwal},
  \bibinfo{person}{Arvind Neelakantan}, \bibinfo{person}{Pranav Shyam},
  \bibinfo{person}{Girish Sastry}, \bibinfo{person}{Amanda Askell},
  \bibinfo{person}{Sandhini Agarwal}, \bibinfo{person}{Ariel Herbert-Voss},
  \bibinfo{person}{Gretchen Krueger}, \bibinfo{person}{Tom Henighan},
  \bibinfo{person}{Rewon Child}, \bibinfo{person}{Aditya Ramesh},
  \bibinfo{person}{Daniel Ziegler}, \bibinfo{person}{Jeffrey Wu},
  \bibinfo{person}{Clemens Winter}, \bibinfo{person}{Chris Hesse},
  \bibinfo{person}{Mark Chen}, \bibinfo{person}{Eric Sigler},
  \bibinfo{person}{Mateusz Litwin}, \bibinfo{person}{Scott Gray},
  \bibinfo{person}{Benjamin Chess}, \bibinfo{person}{Jack Clark},
  \bibinfo{person}{Christopher Berner}, \bibinfo{person}{Sam McCandlish},
  \bibinfo{person}{Alec Radford}, \bibinfo{person}{Ilya Sutskever}, {and}
  \bibinfo{person}{Dario Amodei}.} \bibinfo{year}{2020}\natexlab{}.
\newblock \showarticletitle{Language Models are Few-Shot Learners}. In
  \bibinfo{booktitle}{\emph{Advances in Neural Information Processing
  Systems}}, \bibfield{editor}{\bibinfo{person}{H.~Larochelle},
  \bibinfo{person}{M.~Ranzato}, \bibinfo{person}{R.~Hadsell},
  \bibinfo{person}{M.F. Balcan}, {and} \bibinfo{person}{H.~Lin}} (Eds.),
  Vol.~\bibinfo{volume}{33}. \bibinfo{publisher}{Curran Associates, Inc.},
  \bibinfo{pages}{1877--1901}.
\newblock


\bibitem[Chang et~al\mbox{.}(2024)]%
        {chang-survey}
\bibfield{author}{\bibinfo{person}{Yupeng Chang}, \bibinfo{person}{Xu Wang},
  \bibinfo{person}{Jindong Wang}, \bibinfo{person}{Yuan Wu},
  \bibinfo{person}{Linyi Yang}, \bibinfo{person}{Kaijie Zhu},
  \bibinfo{person}{Hao Chen}, \bibinfo{person}{Xiaoyuan Yi},
  \bibinfo{person}{Cunxiang Wang}, \bibinfo{person}{Yidong Wang},
  \bibinfo{person}{Wei Ye}, \bibinfo{person}{Yue Zhang}, \bibinfo{person}{Yi
  Chang}, \bibinfo{person}{Philip~S. Yu}, \bibinfo{person}{Qiang Yang}, {and}
  \bibinfo{person}{Xing Xie}.} \bibinfo{year}{2024}\natexlab{}.
\newblock \showarticletitle{A Survey on Evaluation of Large Language Models}.
\newblock \bibinfo{journal}{\emph{ACM Trans. Intell. Syst. Technol.}}
  \bibinfo{volume}{15}, \bibinfo{number}{3}, Article \bibinfo{articleno}{39}
  (\bibinfo{date}{March} \bibinfo{year}{2024}), \bibinfo{numpages}{45}~pages.
\newblock
\showISSN{2157-6904}
\href{https://doi.org/10.1145/3641289}{doi:\nolinkurl{10.1145/3641289}}


\bibitem[Chen et~al\mbox{.}(2023)]%
        {chen2023program}
\bibfield{author}{\bibinfo{person}{Wenhu Chen}, \bibinfo{person}{Xueguang Ma},
  \bibinfo{person}{Xinyi Wang}, {and} \bibinfo{person}{William~W. Cohen}.}
  \bibinfo{year}{2023}\natexlab{}.
\newblock \showarticletitle{Program of Thoughts Prompting: Disentangling
  Computation from Reasoning for Numerical Reasoning Tasks}.
\newblock \bibinfo{journal}{\emph{Transactions on Machine Learning Research}}
  (\bibinfo{year}{2023}).
\newblock
\showISSN{2835-8856}
\urldef\tempurl%
\url{https://openreview.net/forum?id=YfZ4ZPt8zd}
\showURL{%
\tempurl}


\bibitem[Cho et~al\mbox{.}(2014)]%
        {gru-rnn}
\bibfield{author}{\bibinfo{person}{Kyunghyun Cho}, \bibinfo{person}{Bart van
  Merri{\"e}nboer}, \bibinfo{person}{Dzmitry Bahdanau}, {and}
  \bibinfo{person}{Yoshua Bengio}.} \bibinfo{year}{2014}\natexlab{}.
\newblock \showarticletitle{On the Properties of Neural Machine Translation:
  Encoder{--}Decoder Approaches}. In \bibinfo{booktitle}{\emph{Proceedings of
  {SSST}-8, Eighth Workshop on Syntax, Semantics and Structure in Statistical
  Translation}}, \bibfield{editor}{\bibinfo{person}{Dekai Wu},
  \bibinfo{person}{Marine Carpuat}, \bibinfo{person}{Xavier Carreras}, {and}
  \bibinfo{person}{Eva~Maria Vecchi}} (Eds.). \bibinfo{publisher}{Association
  for Computational Linguistics}, \bibinfo{address}{Doha, Qatar},
  \bibinfo{pages}{103--111}.
\newblock
\href{https://doi.org/10.3115/v1/W14-4012}{doi:\nolinkurl{10.3115/v1/W14-4012}}


\bibitem[Conneau et~al\mbox{.}(2017)]%
        {conneau-supervised-nlp}
\bibfield{author}{\bibinfo{person}{Alexis Conneau}, \bibinfo{person}{Douwe
  Kiela}, \bibinfo{person}{Holger Schwenk}, \bibinfo{person}{Lo{\"\i}c
  Barrault}, {and} \bibinfo{person}{Antoine Bordes}.}
  \bibinfo{year}{2017}\natexlab{}.
\newblock \showarticletitle{Supervised Learning of Universal Sentence
  Representations from Natural Language Inference Data}. In
  \bibinfo{booktitle}{\emph{Proceedings of the 2017 Conference on Empirical
  Methods in Natural Language Processing}}. \bibinfo{publisher}{Association for
  Computational Linguistics}, \bibinfo{address}{Copenhagen, Denmark},
  \bibinfo{pages}{670--680}.
\newblock
\href{https://doi.org/10.18653/v1/D17-1070}{doi:\nolinkurl{10.18653/v1/D17-1070}}


\bibitem[Corporation(2017)]%
        {volta-v100}
\bibfield{author}{\bibinfo{person}{NVIDIA Corporation}.} \bibinfo{year}{June
  2017}\natexlab{}.
\newblock \bibinfo{booktitle}{\emph{Volta Tesla V100 GPU Architecture
  Whitepaper}}.
\newblock
\urldef\tempurl%
\url{http://images.nvidia.com/content/volta-architecture/pdf/volta-architecture-whitepaper.pdf}
\showURL{%
\tempurl}


\bibitem[Croft and Lafferty(2003)]%
        {croft2003stat}
\bibfield{author}{\bibinfo{person}{Bruce Croft} {and} \bibinfo{person}{John
  Lafferty}.} \bibinfo{year}{2003}\natexlab{}.
\newblock \bibinfo{booktitle}{\emph{Language modeling for information
  retrieval}}. Vol.~\bibinfo{volume}{13}.
\newblock \bibinfo{publisher}{Springer Science \& Business Media}.
\newblock


\bibitem[Dai et~al\mbox{.}(2024)]%
        {dai2024deepseekmoeultimateexpertspecialization}
\bibfield{author}{\bibinfo{person}{Damai Dai}, \bibinfo{person}{Chengqi Deng},
  \bibinfo{person}{Chenggang Zhao}, \bibinfo{person}{R.~X. Xu},
  \bibinfo{person}{Huazuo Gao}, \bibinfo{person}{Deli Chen},
  \bibinfo{person}{Jiashi Li}, \bibinfo{person}{Wangding Zeng},
  \bibinfo{person}{Xingkai Yu}, \bibinfo{person}{Y. Wu},
  \bibinfo{person}{Zhenda Xie}, \bibinfo{person}{Y.~K. Li},
  \bibinfo{person}{Panpan Huang}, \bibinfo{person}{Fuli Luo},
  \bibinfo{person}{Chong Ruan}, \bibinfo{person}{Zhifang Sui}, {and}
  \bibinfo{person}{Wenfeng Liang}.} \bibinfo{year}{2024}\natexlab{}.
\newblock \bibinfo{title}{DeepSeekMoE: Towards Ultimate Expert Specialization
  in Mixture-of-Experts Language Models}.
\newblock
\showeprint[arxiv]{2401.06066}~[cs.CL]
\urldef\tempurl%
\url{https://arxiv.org/abs/2401.06066}
\showURL{%
\tempurl}


\bibitem[Devlin et~al\mbox{.}(2019)]%
        {devlin-bert}
\bibfield{author}{\bibinfo{person}{Jacob Devlin}, \bibinfo{person}{Ming-Wei
  Chang}, \bibinfo{person}{Kenton Lee}, {and} \bibinfo{person}{Kristina
  Toutanova}.} \bibinfo{year}{2019}\natexlab{}.
\newblock \bibinfo{title}{BERT: Pre-training of Deep Bidirectional Transformers
  for Language Understanding}.
\newblock
\showeprint[arxiv]{1810.04805}~[cs.CL]
\urldef\tempurl%
\url{https://arxiv.org/abs/1810.04805}
\showURL{%
\tempurl}


\bibitem[Diamatopoulos et~al\mbox{.}(2024)]%
        {blockchain-diamat}
\bibfield{author}{\bibinfo{person}{Georgios Diamatopoulos},
  \bibinfo{person}{Georgios Theodoropoulos}, \bibinfo{person}{Nikos Tziritas},
  {and} \bibinfo{person}{Rami Bahsoon}.} \bibinfo{year}{2024}\natexlab{}.
\newblock \showarticletitle{Towards LLM Augmented Discrete Event Simulation of
  Blockchain Systems}. In \bibinfo{booktitle}{\emph{Proceedings of the 38th ACM
  SIGSIM Conference on Principles of Advanced Discrete Simulation}} (Atlanta,
  GA, USA) \emph{(\bibinfo{series}{SIGSIM-PADS '24})}.
  \bibinfo{publisher}{Association for Computing Machinery},
  \bibinfo{address}{New York, NY, USA}, \bibinfo{pages}{75–76}.
\newblock
\showISBNx{9798400703638}
\href{https://doi.org/10.1145/3615979.3662156}{doi:\nolinkurl{10.1145/3615979.3662156}}


\bibitem[Ding et~al\mbox{.}(2024)]%
        {ding024everythingthoughts}
\bibfield{author}{\bibinfo{person}{Ruomeng Ding}, \bibinfo{person}{Chaoyun
  Zhang}, \bibinfo{person}{Lu Wang}, \bibinfo{person}{Yong Xu},
  \bibinfo{person}{Minghua Ma}, \bibinfo{person}{Wei Zhang},
  \bibinfo{person}{Si Qin}, \bibinfo{person}{Saravan Rajmohan},
  \bibinfo{person}{Qingwei Lin}, {and} \bibinfo{person}{Dongmei Zhang}.}
  \bibinfo{year}{2024}\natexlab{}.
\newblock \showarticletitle{Everything of Thoughts: Defying the Law of Penrose
  Triangle for Thought Generation}. In \bibinfo{booktitle}{\emph{ACL
  (Findings)}}. \bibinfo{pages}{1638--1662}.
\newblock
\urldef\tempurl%
\url{https://doi.org/10.18653/v1/2024.findings-acl.95}
\showURL{%
\tempurl}


\bibitem[Dubey et~al\mbox{.}(2024)]%
        {dubey2024llama3}
\bibfield{author}{\bibinfo{person}{Abhimanyu Dubey}, \bibinfo{person}{Abhinav
  Jauhri}, \bibinfo{person}{Abhinav Pandey}, \bibinfo{person}{Abhishek Kadian},
  \bibinfo{person}{Ahmad Al-Dahle}, \bibinfo{person}{Aiesha Letman},
  \bibinfo{person}{Akhil Mathur}, \bibinfo{person}{Alan Schelten},
  \bibinfo{person}{Amy Yang}, {and} \bibinfo{person}{Angela~Fan et. al.}}
  \bibinfo{year}{2024}\natexlab{}.
\newblock \bibinfo{title}{The Llama 3 Herd of Models}.
\newblock
\showeprint[arxiv]{2407.21783}~[cs.AI]
\urldef\tempurl%
\url{https://arxiv.org/abs/2407.21783}
\showURL{%
\tempurl}


\bibitem[Eker et~al\mbox{.}(2021)]%
        {eker_load-aware_2021-2}
\bibfield{author}{\bibinfo{person}{Ali Eker}, \bibinfo{person}{Yehia Arafa},
  \bibinfo{person}{Abdel-Hameed~A. Badawy}, \bibinfo{person}{Nandakishore
  Santhi}, \bibinfo{person}{Stephan Eidenbenz}, {and} \bibinfo{person}{Dmitry
  Ponomarev}.} \bibinfo{year}{2021}\natexlab{}.
\newblock \showarticletitle{Load-{Aware} {Dynamic} {Time} {Synchronization} in
  {Parallel} {Discrete} {Event} {Simulation}}. In
  \bibinfo{booktitle}{\emph{Proceedings of the 2021 {ACM} {SIGSIM} {Conference}
  on {Principles} of {Advanced} {Discrete} {Simulation}}}
  \emph{(\bibinfo{series}{{SIGSIM}-{PADS} '21})}.
  \bibinfo{publisher}{Association for Computing Machinery},
  \bibinfo{address}{New York, NY, USA}, \bibinfo{pages}{95--105}.
\newblock
\showISBNx{9781450382960}
\href{https://doi.org/10.1145/3437959.3459249}{doi:\nolinkurl{10.1145/3437959.3459249}}


\bibitem[Fatemi et~al\mbox{.}(2024)]%
        {fatemi2024graph}
\bibfield{author}{\bibinfo{person}{Bahare Fatemi}, \bibinfo{person}{Jonathan
  Halcrow}, {and} \bibinfo{person}{Bryan Perozzi}.}
  \bibinfo{year}{2024}\natexlab{}.
\newblock \showarticletitle{Talk like a Graph: Encoding Graphs for Large
  Language Models}. In \bibinfo{booktitle}{\emph{The Twelfth International
  Conference on Learning Representations}}.
\newblock
\urldef\tempurl%
\url{https://openreview.net/forum?id=IuXR1CCrSi}
\showURL{%
\tempurl}


\bibitem[Fujimoto(1990)]%
        {fujimoto-pdes}
\bibfield{author}{\bibinfo{person}{Richard~M. Fujimoto}.}
  \bibinfo{year}{1990}\natexlab{}.
\newblock \showarticletitle{Parallel discrete event simulation}.
\newblock \bibinfo{journal}{\emph{Commun. ACM}} \bibinfo{volume}{33},
  \bibinfo{number}{10} (\bibinfo{date}{Oct.} \bibinfo{year}{1990}),
  \bibinfo{pages}{30–53}.
\newblock
\showISSN{0001-0782}
\href{https://doi.org/10.1145/84537.84545}{doi:\nolinkurl{10.1145/84537.84545}}


\bibitem[Fujimoto et~al\mbox{.}(2003)]%
        {fujimoto-pdes-network}
\bibfield{author}{\bibinfo{person}{Richard~M. Fujimoto}, \bibinfo{person}{K.
  Perumalla}, \bibinfo{person}{A. Park}, \bibinfo{person}{H. Wu},
  \bibinfo{person}{M.H. Ammar}, {and} \bibinfo{person}{G.F. Riley}.}
  \bibinfo{year}{2003}\natexlab{}.
\newblock \showarticletitle{Large-scale network simulation: how big? how
  fast?}. In \bibinfo{booktitle}{\emph{11th IEEE/ACM International Symposium on
  Modeling, Analysis and Simulation of Computer Telecommunications Systems,
  2003. MASCOTS 2003.}} \bibinfo{pages}{116--123}.
\newblock
\href{https://doi.org/10.1109/MASCOT.2003.1240649}{doi:\nolinkurl{10.1109/MASCOT.2003.1240649}}


\bibitem[Gao and Lin(2004)]%
        {gao-stat-lm}
\bibfield{author}{\bibinfo{person}{Jianfeng Gao} {and}
  \bibinfo{person}{Chin-Yew Lin}.} \bibinfo{year}{2004}\natexlab{}.
\newblock \showarticletitle{Introduction to the special issue on statistical
  language modeling}.
\newblock \bibinfo{journal}{\emph{ACM Transactions on Asian Language
  Information Processing}} \bibinfo{volume}{3}, \bibinfo{number}{2}
  (\bibinfo{date}{June} \bibinfo{year}{2004}), \bibinfo{pages}{87–93}.
\newblock
\showISSN{1530-0226}
\href{https://doi.org/10.1145/1034780.1034781}{doi:\nolinkurl{10.1145/1034780.1034781}}


\bibitem[Gerganov(2024a)]%
        {gguf}
\bibfield{author}{\bibinfo{person}{Georgi Gerganov}.} \bibinfo{year}{Septerber
  2024}\natexlab{a}.
\newblock \bibinfo{booktitle}{\emph{{GGUF}}}.
\newblock
\urldef\tempurl%
\url{https://github.com/ggerganov/ggml/blob/master/docs/gguf.md}
\showURL{%
\tempurl}


\bibitem[Gerganov(2024b)]%
        {llama-cpp}
\bibfield{author}{\bibinfo{person}{Georgi Gerganov}.} \bibinfo{year}{Septerber
  2024}\natexlab{b}.
\newblock \bibinfo{booktitle}{\emph{LLM inference in C/C++}}.
\newblock
\urldef\tempurl%
\url{https://github.com/ggerganov/llama.cpp}
\showURL{%
\tempurl}


\bibitem[Gero et~al\mbox{.}(2023)]%
        {gero2023selfverification}
\bibfield{author}{\bibinfo{person}{Zelalem Gero}, \bibinfo{person}{Chandan
  Singh}, \bibinfo{person}{Hao Cheng}, \bibinfo{person}{Tristan Naumann},
  \bibinfo{person}{Michel Galley}, \bibinfo{person}{Jianfeng Gao}, {and}
  \bibinfo{person}{Hoifung Poon}.} \bibinfo{year}{2023}\natexlab{}.
\newblock \bibinfo{title}{Self-Verification Improves Few-Shot Clinical
  Information Extraction}.
\newblock
\showeprint[arxiv]{2306.00024}~[cs.CL]
\urldef\tempurl%
\url{https://arxiv.org/abs/2306.00024}
\showURL{%
\tempurl}


\bibitem[Giabbanelli(2023)]%
        {gpt-simulation-giabbanelli}
\bibfield{author}{\bibinfo{person}{Philippe~J. Giabbanelli}.}
  \bibinfo{year}{2023}\natexlab{}.
\newblock \showarticletitle{GPT-Based Models Meet Simulation: How to
  Efficiently use Large-Scale Pre-Trained Language Models Across Simulation
  Tasks}. In \bibinfo{booktitle}{\emph{Proceedings of the Winter Simulation
  Conference}} (San Antonio, Texas, USA) \emph{(\bibinfo{series}{WSC '23})}.
  \bibinfo{publisher}{IEEE Press}, \bibinfo{pages}{2920–2931}.
\newblock
\showISBNx{9798350369663}


\bibitem[Hao et~al\mbox{.}(2023)]%
        {hao2023reasoninglanguagemodelplanning}
\bibfield{author}{\bibinfo{person}{Shibo Hao}, \bibinfo{person}{Yi Gu},
  \bibinfo{person}{Haodi Ma}, \bibinfo{person}{Joshua~Jiahua Hong},
  \bibinfo{person}{Zhen Wang}, \bibinfo{person}{Daisy~Zhe Wang}, {and}
  \bibinfo{person}{Zhiting Hu}.} \bibinfo{year}{2023}\natexlab{}.
\newblock \bibinfo{title}{Reasoning with Language Model is Planning with World
  Model}.
\newblock
\showeprint[arxiv]{2305.14992}~[cs.CL]
\urldef\tempurl%
\url{https://arxiv.org/abs/2305.14992}
\showURL{%
\tempurl}


\bibitem[Hernandez-Leal et~al\mbox{.}(2019)]%
        {hernandez2019muti-reinforce}
\bibfield{author}{\bibinfo{person}{Pablo Hernandez-Leal},
  \bibinfo{person}{Bilal Kartal}, {and} \bibinfo{person}{Matthew~E Taylor}.}
  \bibinfo{year}{2019}\natexlab{}.
\newblock \showarticletitle{A survey and critique of multiagent deep
  reinforcement learning}.
\newblock \bibinfo{journal}{\emph{Autonomous Agents and Multi-Agent Systems}}
  \bibinfo{volume}{33}, \bibinfo{number}{6} (\bibinfo{year}{2019}),
  \bibinfo{pages}{750--797}.
\newblock


\bibitem[Hochreiter and Schmidhuber(1997)]%
        {lstm-rnn}
\bibfield{author}{\bibinfo{person}{Sepp Hochreiter} {and}
  \bibinfo{person}{Jürgen Schmidhuber}.} \bibinfo{year}{1997}\natexlab{}.
\newblock \showarticletitle{Long Short-Term Memory}.
\newblock \bibinfo{journal}{\emph{Neural Computation}} \bibinfo{volume}{9},
  \bibinfo{number}{8} (\bibinfo{year}{1997}), \bibinfo{pages}{1735--1780}.
\newblock
\href{https://doi.org/10.1162/neco.1997.9.8.1735}{doi:\nolinkurl{10.1162/neco.1997.9.8.1735}}


\bibitem[Hori et~al\mbox{.}(2018)]%
        {Hori2018speech_rnn}
\bibfield{author}{\bibinfo{person}{Takaaki Hori}, \bibinfo{person}{Jaejin Cho},
  {and} \bibinfo{person}{Shinji Watanabe}.} \bibinfo{year}{2018}\natexlab{}.
\newblock \showarticletitle{End-to-end Speech Recognition With Word-Based Rnn
  Language Models}. In \bibinfo{booktitle}{\emph{2018 IEEE Spoken Language
  Technology Workshop (SLT)}}. \bibinfo{pages}{389--396}.
\newblock
\href{https://doi.org/10.1109/SLT.2018.8639693}{doi:\nolinkurl{10.1109/SLT.2018.8639693}}


\bibitem[Hunter et~al\mbox{.}(2014)]%
        {hunter2014measuring}
\bibfield{author}{\bibinfo{person}{Anthony Hunter}, \bibinfo{person}{Simon
  Parsons}, {and} \bibinfo{person}{Michael Wooldridge}.}
  \bibinfo{year}{2014}\natexlab{}.
\newblock \showarticletitle{Measuring inconsistency in multi-agent systems}.
\newblock \bibinfo{journal}{\emph{KI-K{\"u}nstliche Intelligenz}}
  \bibinfo{volume}{28} (\bibinfo{year}{2014}), \bibinfo{pages}{169--178}.
\newblock


\bibitem[Jiang et~al\mbox{.}(2024)]%
        {jiang2024mixtralexperts}
\bibfield{author}{\bibinfo{person}{Albert~Q. Jiang}, \bibinfo{person}{Alexandre
  Sablayrolles}, \bibinfo{person}{Antoine Roux}, \bibinfo{person}{Arthur
  Mensch}, \bibinfo{person}{Blanche Savary}, \bibinfo{person}{Chris Bamford},
  \bibinfo{person}{Devendra~Singh Chaplot}, \bibinfo{person}{Diego de~las
  Casas}, \bibinfo{person}{Emma~Bou Hanna}, \bibinfo{person}{Florian Bressand},
  \bibinfo{person}{Gianna Lengyel}, \bibinfo{person}{Guillaume Bour},
  \bibinfo{person}{Guillaume Lample}, \bibinfo{person}{Lélio~Renard Lavaud},
  \bibinfo{person}{Lucile Saulnier}, \bibinfo{person}{Marie-Anne Lachaux},
  \bibinfo{person}{Pierre Stock}, \bibinfo{person}{Sandeep Subramanian},
  \bibinfo{person}{Sophia Yang}, \bibinfo{person}{Szymon Antoniak},
  \bibinfo{person}{Teven~Le Scao}, \bibinfo{person}{Théophile Gervet},
  \bibinfo{person}{Thibaut Lavril}, \bibinfo{person}{Thomas Wang},
  \bibinfo{person}{Timothée Lacroix}, {and} \bibinfo{person}{William~El
  Sayed}.} \bibinfo{year}{2024}\natexlab{}.
\newblock \bibinfo{title}{Mixtral of Experts}.
\newblock
\showeprint[arxiv]{2401.04088}~[cs.LG]
\urldef\tempurl%
\url{https://arxiv.org/abs/2401.04088}
\showURL{%
\tempurl}


\bibitem[Khot et~al\mbox{.}(2023)]%
        {khot2023decomposedprompting}
\bibfield{author}{\bibinfo{person}{Tushar Khot}, \bibinfo{person}{Harsh
  Trivedi}, \bibinfo{person}{Matthew Finlayson}, \bibinfo{person}{Yao Fu},
  \bibinfo{person}{Kyle Richardson}, \bibinfo{person}{Peter Clark}, {and}
  \bibinfo{person}{Ashish Sabharwal}.} \bibinfo{year}{2023}\natexlab{}.
\newblock \bibinfo{title}{Decomposed Prompting: A Modular Approach for Solving
  Complex Tasks}.
\newblock
\showeprint[arxiv]{2210.02406}~[cs.CL]
\urldef\tempurl%
\url{https://arxiv.org/abs/2210.02406}
\showURL{%
\tempurl}


\bibitem[Kojima et~al\mbox{.}(2024)]%
        {kojima2024-zero-shot-cot}
\bibfield{author}{\bibinfo{person}{Takeshi Kojima},
  \bibinfo{person}{Shixiang~Shane Gu}, \bibinfo{person}{Machel Reid},
  \bibinfo{person}{Yutaka Matsuo}, {and} \bibinfo{person}{Yusuke Iwasawa}.}
  \bibinfo{year}{2024}\natexlab{}.
\newblock \showarticletitle{Large language models are zero-shot reasoners}. In
  \bibinfo{booktitle}{\emph{Proceedings of the 36th International Conference on
  Neural Information Processing Systems}} (New Orleans, LA, USA)
  \emph{(\bibinfo{series}{NIPS '22})}. \bibinfo{publisher}{Curran Associates
  Inc.}, \bibinfo{address}{Red Hook, NY, USA}, Article
  \bibinfo{articleno}{1613}, \bibinfo{numpages}{15}~pages.
\newblock
\showISBNx{9781713871088}


\bibitem[Kombrink et~al\mbox{.}(2011)]%
        {kombrink2011recurrent}
\bibfield{author}{\bibinfo{person}{Stefan Kombrink}, \bibinfo{person}{Tomas
  Mikolov}, \bibinfo{person}{Martin Karafi{\'a}t}, {and}
  \bibinfo{person}{Luk{\'a}s Burget}.} \bibinfo{year}{2011}\natexlab{}.
\newblock \showarticletitle{Recurrent Neural Network Based Language Modeling in
  Meeting Recognition.}. In \bibinfo{booktitle}{\emph{Interspeech}},
  Vol.~\bibinfo{volume}{11}. \bibinfo{pages}{2877--2880}.
\newblock


\bibitem[Liu et~al\mbox{.}(2024a)]%
        {liu2024tuninglanguagemodelsproxy}
\bibfield{author}{\bibinfo{person}{Alisa Liu}, \bibinfo{person}{Xiaochuang
  Han}, \bibinfo{person}{Yizhong Wang}, \bibinfo{person}{Yulia Tsvetkov},
  \bibinfo{person}{Yejin Choi}, {and} \bibinfo{person}{Noah~A. Smith}.}
  \bibinfo{year}{2024}\natexlab{a}.
\newblock \bibinfo{title}{Tuning Language Models by Proxy}.
\newblock
\showeprint[arxiv]{2401.08565}~[cs.CL]
\urldef\tempurl%
\url{https://arxiv.org/abs/2401.08565}
\showURL{%
\tempurl}


\bibitem[Liu et~al\mbox{.}(2024b)]%
        {liu2024llmatransportation}
\bibfield{author}{\bibinfo{person}{Tianming Liu}, \bibinfo{person}{Jirong
  Yang}, {and} \bibinfo{person}{Yafeng Yin}.} \bibinfo{year}{2024}\natexlab{b}.
\newblock \bibinfo{title}{Toward LLM-Agent-Based Modeling of Transportation
  Systems: A Conceptual Framework}.
\newblock
\showeprint[arxiv]{2412.06681}~[cs.AI]
\urldef\tempurl%
\url{https://arxiv.org/abs/2412.06681}
\showURL{%
\tempurl}


\bibitem[Liu and Croft(2005)]%
        {liu2005statistical}
\bibfield{author}{\bibinfo{person}{Xiaoyong Liu} {and} \bibinfo{person}{W~Bruce
  Croft}.} \bibinfo{year}{2005}\natexlab{}.
\newblock \showarticletitle{Statistical language modeling for information
  retrieval.}
\newblock \bibinfo{journal}{\emph{Annu. Rev. Inf. Sci. Technol.}}
  \bibinfo{volume}{39}, \bibinfo{number}{1} (\bibinfo{year}{2005}),
  \bibinfo{pages}{1--31}.
\newblock


\bibitem[Mikolov et~al\mbox{.}(2010)]%
        {mikolov2010recurrent}
\bibfield{author}{\bibinfo{person}{Tomas Mikolov}, \bibinfo{person}{Martin
  Karafi{\'a}t}, \bibinfo{person}{Lukas Burget}, \bibinfo{person}{Jan
  Cernock{\`y}}, {and} \bibinfo{person}{Sanjeev Khudanpur}.}
  \bibinfo{year}{2010}\natexlab{}.
\newblock \showarticletitle{Recurrent neural network based language model.}. In
  \bibinfo{booktitle}{\emph{Interspeech}}, Vol.~\bibinfo{volume}{2}. Makuhari,
  \bibinfo{pages}{1045--1048}.
\newblock


\bibitem[Ming(2023)]%
        {ming2023exploration}
\bibfield{author}{\bibinfo{person}{Guo Ming}.} \bibinfo{year}{2023}\natexlab{}.
\newblock \showarticletitle{Exploration of the intelligent control system of
  autonomous vehicles based on edge computing}.
\newblock \bibinfo{journal}{\emph{PLoS One}} \bibinfo{volume}{18},
  \bibinfo{number}{2} (\bibinfo{year}{2023}), \bibinfo{pages}{e0281294}.
\newblock


\bibitem[Muqeeth et~al\mbox{.}(2024)]%
        {muqeeth2024learningroutespecializedexperts}
\bibfield{author}{\bibinfo{person}{Mohammed Muqeeth}, \bibinfo{person}{Haokun
  Liu}, \bibinfo{person}{Yufan Liu}, {and} \bibinfo{person}{Colin Raffel}.}
  \bibinfo{year}{2024}\natexlab{}.
\newblock \bibinfo{title}{Learning to Route Among Specialized Experts for
  Zero-Shot Generalization}.
\newblock
\showeprint[arxiv]{2402.05859}~[cs.LG]
\urldef\tempurl%
\url{https://arxiv.org/abs/2402.05859}
\showURL{%
\tempurl}


\bibitem[OpenAI et~al\mbox{.}(2024)]%
        {openai2024gpt4}
\bibfield{author}{\bibinfo{person}{OpenAI}, \bibinfo{person}{Josh Achiam},
  \bibinfo{person}{Steven Adler}, \bibinfo{person}{Sandhini Agarwal},
  \bibinfo{person}{Lama Ahmad}, \bibinfo{person}{Ilge Akkaya},
  \bibinfo{person}{Florencia~Leoni Aleman}, \bibinfo{person}{Diogo Almeida},
  \bibinfo{person}{Janko Altenschmidt}, {and} \bibinfo{person}{Sam~Altman et.
  al.}} \bibinfo{year}{2024}\natexlab{}.
\newblock \bibinfo{title}{GPT-4 Technical Report}.
\newblock
\showeprint[arxiv]{2303.08774}~[cs.CL]
\urldef\tempurl%
\url{https://arxiv.org/abs/2303.08774}
\showURL{%
\tempurl}


\bibitem[Perumalla(2007)]%
        {perumalla-wrap-bg}
\bibfield{author}{\bibinfo{person}{Kalyan~S. Perumalla}.}
  \bibinfo{year}{2007}\natexlab{}.
\newblock \showarticletitle{Scaling time warp-based discrete event execution to
  104 processors on a Blue Gene supercomputer}. In
  \bibinfo{booktitle}{\emph{Proceedings of the 4th International Conference on
  Computing Frontiers}} (Ischia, Italy) \emph{(\bibinfo{series}{CF '07})}.
  \bibinfo{publisher}{Association for Computing Machinery},
  \bibinfo{address}{New York, NY, USA}, \bibinfo{pages}{69–76}.
\newblock
\showISBNx{9781595936837}
\href{https://doi.org/10.1145/1242531.1242543}{doi:\nolinkurl{10.1145/1242531.1242543}}


\bibitem[Qi et~al\mbox{.}(2024)]%
        {qi2024mutualreasoningslm}
\bibfield{author}{\bibinfo{person}{Zhenting Qi}, \bibinfo{person}{Mingyuan Ma},
  \bibinfo{person}{Jiahang Xu}, \bibinfo{person}{Li~Lyna Zhang},
  \bibinfo{person}{Fan Yang}, {and} \bibinfo{person}{Mao Yang}.}
  \bibinfo{year}{2024}\natexlab{}.
\newblock \bibinfo{title}{Mutual Reasoning Makes Smaller LLMs Stronger
  Problem-Solvers}.
\newblock
\showeprint[arxiv]{2408.06195}~[cs.CL]
\urldef\tempurl%
\url{https://arxiv.org/abs/2408.06195}
\showURL{%
\tempurl}


\bibitem[Qwen et~al\mbox{.}(2025)]%
        {qwen2025}
\bibfield{author}{\bibinfo{person}{Qwen}, \bibinfo{person}{:},
  \bibinfo{person}{An Yang}, \bibinfo{person}{Baosong Yang},
  \bibinfo{person}{Beichen Zhang}, \bibinfo{person}{Binyuan Hui},
  \bibinfo{person}{Bo Zheng}, \bibinfo{person}{Bowen Yu},
  \bibinfo{person}{Chengyuan Li}, \bibinfo{person}{Dayiheng Liu},
  \bibinfo{person}{Fei Huang}, \bibinfo{person}{Haoran Wei},
  \bibinfo{person}{Huan Lin}, \bibinfo{person}{Jian Yang},
  \bibinfo{person}{Jianhong Tu}, \bibinfo{person}{Jianwei Zhang},
  \bibinfo{person}{Jianxin Yang}, \bibinfo{person}{Jiaxi Yang},
  \bibinfo{person}{Jingren Zhou}, \bibinfo{person}{Junyang Lin},
  \bibinfo{person}{Kai Dang}, \bibinfo{person}{Keming Lu},
  \bibinfo{person}{Keqin Bao}, \bibinfo{person}{Kexin Yang},
  \bibinfo{person}{Le Yu}, \bibinfo{person}{Mei Li}, \bibinfo{person}{Mingfeng
  Xue}, \bibinfo{person}{Pei Zhang}, \bibinfo{person}{Qin Zhu},
  \bibinfo{person}{Rui Men}, \bibinfo{person}{Runji Lin},
  \bibinfo{person}{Tianhao Li}, \bibinfo{person}{Tianyi Tang},
  \bibinfo{person}{Tingyu Xia}, \bibinfo{person}{Xingzhang Ren},
  \bibinfo{person}{Xuancheng Ren}, \bibinfo{person}{Yang Fan},
  \bibinfo{person}{Yang Su}, \bibinfo{person}{Yichang Zhang},
  \bibinfo{person}{Yu Wan}, \bibinfo{person}{Yuqiong Liu},
  \bibinfo{person}{Zeyu Cui}, \bibinfo{person}{Zhenru Zhang}, {and}
  \bibinfo{person}{Zihan Qiu}.} \bibinfo{year}{2025}\natexlab{}.
\newblock \bibinfo{title}{Qwen2.5 Technical Report}.
\newblock
\showeprint[arxiv]{2412.15115}~[cs.CL]
\urldef\tempurl%
\url{https://arxiv.org/abs/2412.15115}
\showURL{%
\tempurl}


\bibitem[Radford et~al\mbox{.}(2018)]%
        {radford2018gpt}
\bibfield{author}{\bibinfo{person}{Alec Radford}, \bibinfo{person}{Karthik
  Narasimhan}, \bibinfo{person}{Tim Salimans}, {and} \bibinfo{person}{Ilya
  Sutskever}.} \bibinfo{year}{2018}\natexlab{}.
\newblock \bibinfo{title}{Improving language understanding by generative
  pre-training}.
\newblock
\urldef\tempurl%
\url{https://cdn.openai.com/research-covers/language-unsupervised/language\_understanding\_paper.pdf}
\showURL{%
\tempurl}


\bibitem[Rong et~al\mbox{.}(2014)]%
        {rong-miniSSF}
\bibfield{author}{\bibinfo{person}{Rong Rong}, \bibinfo{person}{Jiang Hao},
  {and} \bibinfo{person}{Jason Liu}.} \bibinfo{year}{2014}\natexlab{}.
\newblock \showarticletitle{Performance Study of a Minimalistic Simulator on
  XSEDE Massively Parallel Systems}. In \bibinfo{booktitle}{\emph{Proceedings
  of the 2014 Annual Conference on Extreme Science and Engineering Discovery
  Environment}} (Atlanta, GA, USA) \emph{(\bibinfo{series}{XSEDE '14})}.
  \bibinfo{publisher}{Association for Computing Machinery},
  \bibinfo{address}{New York, NY, USA}, Article \bibinfo{articleno}{15},
  \bibinfo{numpages}{8}~pages.
\newblock
\showISBNx{9781450328937}
\href{https://doi.org/10.1145/2616498.2616512}{doi:\nolinkurl{10.1145/2616498.2616512}}


\bibitem[Santhi et~al\mbox{.}(2015)]%
        {simian}
\bibfield{author}{\bibinfo{person}{Nandakishore Santhi},
  \bibinfo{person}{Stephan Eidenbenz}, {and} \bibinfo{person}{Jason Liu}.}
  \bibinfo{year}{2015}\natexlab{}.
\newblock \showarticletitle{The Simian concept: Parallel Discrete Event
  Simulation with interpreted languages and just-in-time compilation}. In
  \bibinfo{booktitle}{\emph{2015 Winter Simulation Conference (WSC)}}.
  \bibinfo{pages}{3013--3024}.
\newblock
\href{https://doi.org/10.1109/WSC.2015.7408405}{doi:\nolinkurl{10.1109/WSC.2015.7408405}}


\bibitem[Satpute et~al\mbox{.}(2024)]%
        {llm-math-ankit}
\bibfield{author}{\bibinfo{person}{Ankit Satpute}, \bibinfo{person}{Noah
  Gie\ss{}ing}, \bibinfo{person}{Andr\'{e} Greiner-Petter},
  \bibinfo{person}{Moritz Schubotz}, \bibinfo{person}{Olaf Teschke},
  \bibinfo{person}{Akiko Aizawa}, {and} \bibinfo{person}{Bela Gipp}.}
  \bibinfo{year}{2024}\natexlab{}.
\newblock \showarticletitle{Can LLMs Master Math? Investigating Large Language
  Models on Math Stack Exchange}. In \bibinfo{booktitle}{\emph{Proceedings of
  the 47th International ACM SIGIR Conference on Research and Development in
  Information Retrieval}} (Washington DC, USA) \emph{(\bibinfo{series}{SIGIR
  '24})}. \bibinfo{publisher}{Association for Computing Machinery},
  \bibinfo{address}{New York, NY, USA}, \bibinfo{pages}{2316–2320}.
\newblock
\showISBNx{9798400704314}
\href{https://doi.org/10.1145/3626772.3657945}{doi:\nolinkurl{10.1145/3626772.3657945}}


\bibitem[Shazeer et~al\mbox{.}(2017)]%
        {shazeer2017}
\bibfield{author}{\bibinfo{person}{Noam Shazeer}, \bibinfo{person}{*Azalia
  Mirhoseini}, \bibinfo{person}{*Krzysztof Maziarz}, \bibinfo{person}{Andy
  Davis}, \bibinfo{person}{Quoc Le}, \bibinfo{person}{Geoffrey Hinton}, {and}
  \bibinfo{person}{Jeff Dean}.} \bibinfo{year}{2017}\natexlab{}.
\newblock \showarticletitle{Outrageously Large Neural Networks: The
  Sparsely-Gated Mixture-of-Experts Layer}. In
  \bibinfo{booktitle}{\emph{International Conference on Learning
  Representations}}.
\newblock
\urldef\tempurl%
\url{https://openreview.net/forum?id=B1ckMDqlg}
\showURL{%
\tempurl}


\bibitem[Shen et~al\mbox{.}(2024)]%
        {shen2024collab_llms}
\bibfield{author}{\bibinfo{person}{Shannon~Zejiang Shen},
  \bibinfo{person}{Hunter Lang}, \bibinfo{person}{Bailin Wang},
  \bibinfo{person}{Yoon Kim}, {and} \bibinfo{person}{David Sontag}.}
  \bibinfo{year}{2024}\natexlab{}.
\newblock \bibinfo{title}{Learning to Decode Collaboratively with Multiple
  Language Models}.
\newblock
\showeprint[arxiv]{2403.03870}~[cs.CL]
\urldef\tempurl%
\url{https://arxiv.org/abs/2403.03870}
\showURL{%
\tempurl}


\bibitem[Shinn et~al\mbox{.}(2023)]%
        {shinn2023reflexion}
\bibfield{author}{\bibinfo{person}{Noah Shinn}, \bibinfo{person}{Federico
  Cassano}, \bibinfo{person}{Edward Berman}, \bibinfo{person}{Ashwin Gopinath},
  \bibinfo{person}{Karthik Narasimhan}, {and} \bibinfo{person}{Shunyu Yao}.}
  \bibinfo{year}{2023}\natexlab{}.
\newblock \bibinfo{title}{Reflexion: Language Agents with Verbal Reinforcement
  Learning}.
\newblock
\showeprint[arxiv]{2303.11366}~[cs.AI]
\urldef\tempurl%
\url{https://arxiv.org/abs/2303.11366}
\showURL{%
\tempurl}


\bibitem[Singh(2024)]%
        {llm-bad-math}
\bibfield{author}{\bibinfo{person}{Viraj Singh}.} \bibinfo{year}{July 16,
  2024}\natexlab{}.
\newblock \bibinfo{booktitle}{\emph{Why LLMs Are Bad at Math — and How They
  Can Be Better}}.
\newblock
\urldef\tempurl%
\url{https://www.reachcapital.com/2024/07/16/why-llms-are-bad-at-math-and-how-they-can-be-better/}
\showURL{%
\tempurl}


\bibitem[Stone and Veloso(2000)]%
        {stone2000multiagent}
\bibfield{author}{\bibinfo{person}{Peter Stone} {and} \bibinfo{person}{Manuela
  Veloso}.} \bibinfo{year}{2000}\natexlab{}.
\newblock \showarticletitle{Multiagent systems: A survey from a machine
  learning perspective}.
\newblock \bibinfo{journal}{\emph{Autonomous Robots}}  \bibinfo{volume}{8}
  (\bibinfo{year}{2000}), \bibinfo{pages}{345--383}.
\newblock


\bibitem[Strati et~al\mbox{.}(2024)]%
        {strati-gpu-shortage-cross-region}
\bibfield{author}{\bibinfo{person}{Foteini Strati}, \bibinfo{person}{Paul
  Elvinger}, \bibinfo{person}{Tolga Kerimoglu}, {and} \bibinfo{person}{Ana
  Klimovic}.} \bibinfo{year}{2024}\natexlab{}.
\newblock \showarticletitle{ML Training with Cloud GPU Shortages: Is
  Cross-Region the Answer?}. In \bibinfo{booktitle}{\emph{Proceedings of the
  4th Workshop on Machine Learning and Systems}} (Athens, Greece)
  \emph{(\bibinfo{series}{EuroMLSys '24})}. \bibinfo{publisher}{Association for
  Computing Machinery}, \bibinfo{address}{New York, NY, USA},
  \bibinfo{pages}{107–116}.
\newblock
\showISBNx{9798400705410}
\href{https://doi.org/10.1145/3642970.3655843}{doi:\nolinkurl{10.1145/3642970.3655843}}


\bibitem[team(2024)]%
        {mistral-nemo}
\bibfield{author}{\bibinfo{person}{Mistral~AI team}.} \bibinfo{year}{July
  2024}\natexlab{}.
\newblock \bibinfo{booktitle}{\emph{Mistral Nemo}}.
\newblock
\urldef\tempurl%
\url{https://mistral.ai/news/mistral-nemo/}
\showURL{%
\tempurl}


\bibitem[TechPowerUp(2017)]%
        {v100}
\bibfield{author}{\bibinfo{person}{TechPowerUp}.} \bibinfo{year}{June
  2017}\natexlab{}.
\newblock \bibinfo{booktitle}{\emph{NVIDIA Tesla V100 Specs}}.
\newblock
\urldef\tempurl%
\url{https://www.techpowerup.com/gpu-specs/tesla-v100-pcie-16-gb.c2957}
\showURL{%
\tempurl}


\bibitem[Thulasidasan et~al\mbox{.}(2014)]%
        {sunil-simx}
\bibfield{author}{\bibinfo{person}{Sunil Thulasidasan}, \bibinfo{person}{Lukas
  Kroc}, {and} \bibinfo{person}{Stephan Eidenbenz}.}
  \bibinfo{year}{2014}\natexlab{}.
\newblock \showarticletitle{Developing parallel, discrete event simulations in
  Python - first results and user experiences with the SimX library}. In
  \bibinfo{booktitle}{\emph{2014 4th International Conference On Simulation And
  Modeling Methodologies, Technologies And Applications (SIMULTECH)}}.
  \bibinfo{pages}{188--194}.
\newblock
\href{https://doi.org/10.5220/0005042701880194}{doi:\nolinkurl{10.5220/0005042701880194}}


\bibitem[Vaswani et~al\mbox{.}(2017)]%
        {vaswani-transformer}
\bibfield{author}{\bibinfo{person}{Ashish Vaswani}, \bibinfo{person}{Noam
  Shazeer}, \bibinfo{person}{Niki Parmar}, \bibinfo{person}{Jakob Uszkoreit},
  \bibinfo{person}{Llion Jones}, \bibinfo{person}{Aidan~N Gomez},
  \bibinfo{person}{Lukasz Kaiser}, {and} \bibinfo{person}{Illia Polosukhin}.}
  \bibinfo{year}{2017}\natexlab{}.
\newblock \showarticletitle{Attention is All you Need}. In
  \bibinfo{booktitle}{\emph{Advances in Neural Information Processing
  Systems}}, \bibfield{editor}{\bibinfo{person}{I.~Guyon},
  \bibinfo{person}{U.~Von Luxburg}, \bibinfo{person}{S.~Bengio},
  \bibinfo{person}{H.~Wallach}, \bibinfo{person}{R.~Fergus},
  \bibinfo{person}{S.~Vishwanathan}, {and} \bibinfo{person}{R.~Garnett}}
  (Eds.), Vol.~\bibinfo{volume}{30}. \bibinfo{publisher}{Curran Associates,
  Inc.}
\newblock


\bibitem[Wang et~al\mbox{.}(2023)]%
        {wang2023dynamicreflection}
\bibfield{author}{\bibinfo{person}{Yu Wang}, \bibinfo{person}{Zhiwei Liu},
  \bibinfo{person}{Jianguo Zhang}, \bibinfo{person}{Weiran Yao},
  \bibinfo{person}{Shelby Heinecke}, {and} \bibinfo{person}{Philip~S. Yu}.}
  \bibinfo{year}{2023}\natexlab{}.
\newblock \bibinfo{title}{DRDT: Dynamic Reflection with Divergent Thinking for
  LLM-based Sequential Recommendation}.
\newblock
\showeprint[arxiv]{2312.11336}~[cs.IR]
\urldef\tempurl%
\url{https://arxiv.org/abs/2312.11336}
\showURL{%
\tempurl}


\bibitem[Wei et~al\mbox{.}(2022a)]%
        {wei2022finetuned-zero-shot}
\bibfield{author}{\bibinfo{person}{Jason Wei}, \bibinfo{person}{Maarten Bosma},
  \bibinfo{person}{Vincent Zhao}, \bibinfo{person}{Kelvin Guu},
  \bibinfo{person}{Adams~Wei Yu}, \bibinfo{person}{Brian Lester},
  \bibinfo{person}{Nan Du}, \bibinfo{person}{Andrew~M. Dai}, {and}
  \bibinfo{person}{Quoc~V Le}.} \bibinfo{year}{2022}\natexlab{a}.
\newblock \showarticletitle{Finetuned Language Models are Zero-Shot Learners}.
  In \bibinfo{booktitle}{\emph{International Conference on Learning
  Representations}}.
\newblock
\urldef\tempurl%
\url{https://openreview.net/forum?id=gEZrGCozdqR}
\showURL{%
\tempurl}


\bibitem[Wei et~al\mbox{.}(2022b)]%
        {wei2022emergent_llm}
\bibfield{author}{\bibinfo{person}{Jason Wei}, \bibinfo{person}{Yi Tay},
  \bibinfo{person}{Rishi Bommasani}, \bibinfo{person}{Colin Raffel},
  \bibinfo{person}{Barret Zoph}, \bibinfo{person}{Sebastian Borgeaud},
  \bibinfo{person}{Dani Yogatama}, \bibinfo{person}{Maarten Bosma},
  \bibinfo{person}{Denny Zhou}, \bibinfo{person}{Donald Metzler},
  \bibinfo{person}{Ed~H. Chi}, \bibinfo{person}{Tatsunori Hashimoto},
  \bibinfo{person}{Oriol Vinyals}, \bibinfo{person}{Percy Liang},
  \bibinfo{person}{Jeff Dean}, {and} \bibinfo{person}{William Fedus}.}
  \bibinfo{year}{2022}\natexlab{b}.
\newblock \showarticletitle{Emergent Abilities of Large Language Models}.
\newblock \bibinfo{journal}{\emph{Transactions on Machine Learning Research}}
  (\bibinfo{year}{2022}).
\newblock
\showISSN{2835-8856}
\urldef\tempurl%
\url{https://openreview.net/forum?id=yzkSU5zdwD}
\showURL{%
\tempurl}
\newblock
\shownote{Survey Certification}.


\bibitem[Wei et~al\mbox{.}(2024)]%
        {wei-cot}
\bibfield{author}{\bibinfo{person}{Jason Wei}, \bibinfo{person}{Xuezhi Wang},
  \bibinfo{person}{Dale Schuurmans}, \bibinfo{person}{Maarten Bosma},
  \bibinfo{person}{Brian Ichter}, \bibinfo{person}{Fei Xia},
  \bibinfo{person}{Ed~H. Chi}, \bibinfo{person}{Quoc~V. Le}, {and}
  \bibinfo{person}{Denny Zhou}.} \bibinfo{year}{2024}\natexlab{}.
\newblock \showarticletitle{Chain-of-thought prompting elicits reasoning in
  large language models}. In \bibinfo{booktitle}{\emph{Proceedings of the 36th
  International Conference on Neural Information Processing Systems}} (New
  Orleans, LA, USA) \emph{(\bibinfo{series}{NIPS '22})}.
  \bibinfo{publisher}{Curran Associates Inc.}, \bibinfo{address}{Red Hook, NY,
  USA}, Article \bibinfo{articleno}{1800}, \bibinfo{numpages}{14}~pages.
\newblock
\showISBNx{9781713871088}


\bibitem[Wooldridge(2009)]%
        {wooldridge2009multiagent}
\bibfield{author}{\bibinfo{person}{Michael Wooldridge}.}
  \bibinfo{year}{2009}\natexlab{}.
\newblock \bibinfo{booktitle}{\emph{An introduction to multiagent systems}}.
\newblock \bibinfo{publisher}{John wiley \& sons}.
\newblock


\bibitem[Wu et~al\mbox{.}(2023)]%
        {wu2023smartagentbasedmodelinguse}
\bibfield{author}{\bibinfo{person}{Zengqing Wu}, \bibinfo{person}{Run Peng},
  \bibinfo{person}{Xu Han}, \bibinfo{person}{Shuyuan Zheng},
  \bibinfo{person}{Yixin Zhang}, {and} \bibinfo{person}{Chuan Xiao}.}
  \bibinfo{year}{2023}\natexlab{}.
\newblock \bibinfo{title}{Smart Agent-Based Modeling: On the Use of Large
  Language Models in Computer Simulations}.
\newblock
\showeprint[arxiv]{2311.06330}~[cs.AI]
\urldef\tempurl%
\url{https://arxiv.org/abs/2311.06330}
\showURL{%
\tempurl}


\bibitem[Xue et~al\mbox{.}(2024)]%
        {xue2024openmoeearlyeffortopen}
\bibfield{author}{\bibinfo{person}{Fuzhao Xue}, \bibinfo{person}{Zian Zheng},
  \bibinfo{person}{Yao Fu}, \bibinfo{person}{Jinjie Ni},
  \bibinfo{person}{Zangwei Zheng}, \bibinfo{person}{Wangchunshu Zhou}, {and}
  \bibinfo{person}{Yang You}.} \bibinfo{year}{2024}\natexlab{}.
\newblock \bibinfo{title}{OpenMoE: An Early Effort on Open Mixture-of-Experts
  Language Models}.
\newblock
\showeprint[arxiv]{2402.01739}~[cs.CL]
\urldef\tempurl%
\url{https://arxiv.org/abs/2402.01739}
\showURL{%
\tempurl}


\bibitem[Yao et~al\mbox{.}(2023)]%
        {yao2023treethoughts}
\bibfield{author}{\bibinfo{person}{Shunyu Yao}, \bibinfo{person}{Dian Yu},
  \bibinfo{person}{Jeffrey Zhao}, \bibinfo{person}{Izhak Shafran},
  \bibinfo{person}{Thomas~L. Griffiths}, \bibinfo{person}{Yuan Cao}, {and}
  \bibinfo{person}{Karthik Narasimhan}.} \bibinfo{year}{2023}\natexlab{}.
\newblock \bibinfo{title}{Tree of Thoughts: Deliberate Problem Solving with
  Large Language Models}.
\newblock
\showeprint[arxiv]{2305.10601}~[cs.CL]
\urldef\tempurl%
\url{https://arxiv.org/abs/2305.10601}
\showURL{%
\tempurl}


\bibitem[Zhang et~al\mbox{.}(2024a)]%
        {zhang2024accessinggpt4levelmathematical}
\bibfield{author}{\bibinfo{person}{Di Zhang}, \bibinfo{person}{Xiaoshui Huang},
  \bibinfo{person}{Dongzhan Zhou}, \bibinfo{person}{Yuqiang Li}, {and}
  \bibinfo{person}{Wanli Ouyang}.} \bibinfo{year}{2024}\natexlab{a}.
\newblock \bibinfo{title}{Accessing GPT-4 level Mathematical Olympiad Solutions
  via Monte Carlo Tree Self-refine with LLaMa-3 8B}.
\newblock
\showeprint[arxiv]{2406.07394}~[cs.AI]
\urldef\tempurl%
\url{https://arxiv.org/abs/2406.07394}
\showURL{%
\tempurl}


\bibitem[Zhang et~al\mbox{.}(2024b)]%
        {zhang2024metaprompting}
\bibfield{author}{\bibinfo{person}{Yifan Zhang}, \bibinfo{person}{Yang Yuan},
  {and} \bibinfo{person}{Andrew Chi-Chih Yao}.}
  \bibinfo{year}{2024}\natexlab{b}.
\newblock \bibinfo{title}{Meta Prompting for AI Systems}.
\newblock
\showeprint[arxiv]{2311.11482}~[cs.AI]
\urldef\tempurl%
\url{https://arxiv.org/abs/2311.11482}
\showURL{%
\tempurl}


\bibitem[Zheng et~al\mbox{.}(2024)]%
        {zheng2024abstraction}
\bibfield{author}{\bibinfo{person}{Huaixiu~Steven Zheng},
  \bibinfo{person}{Swaroop Mishra}, \bibinfo{person}{Xinyun Chen},
  \bibinfo{person}{Heng-Tze Cheng}, \bibinfo{person}{Ed~H. Chi},
  \bibinfo{person}{Quoc~V Le}, {and} \bibinfo{person}{Denny Zhou}.}
  \bibinfo{year}{2024}\natexlab{}.
\newblock \showarticletitle{Take a Step Back: Evoking Reasoning via Abstraction
  in Large Language Models}. In \bibinfo{booktitle}{\emph{The Twelfth
  International Conference on Learning Representations}}.
\newblock
\urldef\tempurl%
\url{https://openreview.net/forum?id=3bq3jsvcQ1}
\showURL{%
\tempurl}


\bibitem[Zhou et~al\mbox{.}(2024)]%
        {zhou2024solving}
\bibfield{author}{\bibinfo{person}{Aojun Zhou}, \bibinfo{person}{Ke Wang},
  \bibinfo{person}{Zimu Lu}, \bibinfo{person}{Weikang Shi},
  \bibinfo{person}{Sichun Luo}, \bibinfo{person}{Zipeng Qin},
  \bibinfo{person}{Shaoqing Lu}, \bibinfo{person}{Anya Jia},
  \bibinfo{person}{Linqi Song}, \bibinfo{person}{Mingjie Zhan}, {and}
  \bibinfo{person}{Hongsheng Li}.} \bibinfo{year}{2024}\natexlab{}.
\newblock \showarticletitle{Solving Challenging Math Word Problems Using
  {GPT}-4 Code Interpreter with Code-based Self-Verification}. In
  \bibinfo{booktitle}{\emph{The Twelfth International Conference on Learning
  Representations}}.
\newblock
\urldef\tempurl%
\url{https://openreview.net/forum?id=c8McWs4Av0}
\showURL{%
\tempurl}


\bibitem[Zhou et~al\mbox{.}(2023)]%
        {zhou2023leasttomost}
\bibfield{author}{\bibinfo{person}{Denny Zhou}, \bibinfo{person}{Nathanael
  Sch{\"a}rli}, \bibinfo{person}{Le Hou}, \bibinfo{person}{Jason Wei},
  \bibinfo{person}{Nathan Scales}, \bibinfo{person}{Xuezhi Wang},
  \bibinfo{person}{Dale Schuurmans}, \bibinfo{person}{Claire Cui},
  \bibinfo{person}{Olivier Bousquet}, \bibinfo{person}{Quoc~V Le}, {and}
  \bibinfo{person}{Ed~H. Chi}.} \bibinfo{year}{2023}\natexlab{}.
\newblock \showarticletitle{Least-to-Most Prompting Enables Complex Reasoning
  in Large Language Models}. In \bibinfo{booktitle}{\emph{The Eleventh
  International Conference on Learning Representations}}.
\newblock
\urldef\tempurl%
\url{https://openreview.net/forum?id=WZH7099tgfM}
\showURL{%
\tempurl}


\bibitem[Zhou et~al\mbox{.}(2022)]%
        {zhou2022mixture}
\bibfield{author}{\bibinfo{person}{Yanqi Zhou}, \bibinfo{person}{Tao Lei},
  \bibinfo{person}{Hanxiao Liu}, \bibinfo{person}{Nan Du},
  \bibinfo{person}{Yanping Huang}, \bibinfo{person}{Vincent Zhao},
  \bibinfo{person}{Andrew~M Dai}, \bibinfo{person}{Quoc~V Le},
  \bibinfo{person}{James Laudon}, {et~al\mbox{.}}}
  \bibinfo{year}{2022}\natexlab{}.
\newblock \showarticletitle{Mixture-of-experts with expert choice routing}.
\newblock \bibinfo{journal}{\emph{Advances in Neural Information Processing
  Systems}}  \bibinfo{volume}{35} (\bibinfo{year}{2022}),
  \bibinfo{pages}{7103--7114}.
\newblock


\end{thebibliography}

% \appendix




\end{document}